\documentclass{article}% Standard Nature Portfolio Reference Style
\usepackage[a4paper, total={7in, 10in}]{geometry}

\usepackage[mathscr]{eucal}
\usepackage{amsthm}
\usepackage{graphicx}
\usepackage{mathtools}
\usepackage[hyphens]{url} 
\usepackage[super]{nth}
\usepackage{authblk}

\usepackage{multirow}
\usepackage{xcolor,colortbl}
\usepackage{amsmath}
\usepackage{amsfonts}

\usepackage{algcompatible}
\usepackage{algorithm}
\usepackage{xspace}

\usepackage[norule, flushmargin]{footmisc}

\usepackage{framed}
\usepackage{booktabs}
\usepackage{hyperref,microtype,subcaption}
\usepackage[onehalfspacing]{setspace}
\usepackage{dcolumn}
\usepackage[normalem]{ulem} 

\newcommand{\calvar}[1]{\ensuremath{\mathcal{#1}}}
\newcommand{\calA}{\calvar{A}}

\newcommand{\calN}{\calvar{N}}

\newcommand{\Select}{\ensuremath{\textsc{\textcolor{violet}{Select}}}}

\newcommand{\Explain}{\ensuremath{\textsc{\textcolor{blue}{Explain}}}}
\newcommand{\Obtain}{\ensuremath{\textsc{\textcolor{olive}{Obtain}}}}
\newcommand{\Revise}{\ensuremath{\textsc{\textcolor{orange}{Revise}}}}

\newcommand{\cf}{\emph{cf.}~}

\newcommand{\etal}{\emph{et al.}~}

\newcommand{\wrmeasure}{\textsc{wr \xspace}}
\newcommand{\wrnospace}{\textsc{wr}}

\title{A Typology for Exploring the Mitigation of Shortcut Behavior}
\date{}

\author[1,2,*]{Felix Friedrich}
\author[1,2]{Wolfgang Stammer}
\author[1,2,3,4]{Patrick Schramowski}
\author[1,2,4,5]{Kristian Kersting}
\affil[1]{Technical University of Darmstadt, Computer Science Department, Artificial Intelligence and Machine Learning Lab, Darmstadt, Germany}
\affil[2]{Hessian Center for Artificial Intelligence (hessian.AI), Darmstadt, Germany}
\affil[3]{LAION, Germany}
\affil[4]{German Center for Artificial Intelligence (DFKI), Germany}
\affil[5]{Technical University of Darmstadt, Centre for Cognitive Science, Darmstadt, Germany}
\affil[*]{Corresponding author: Felix Friedrich (friedrich@cs.tu-darmstadt.de)}

\begin{document}

\maketitle

\begin{abstract}
\noindent As machine learning models become larger, and are increasingly trained on large and uncurated data sets in weakly supervised mode, it becomes important to establish mechanisms for inspecting, interacting, and revising models. These are necessary to mitigate shortcut learning effects and to guarantee that the model's learned knowledge is aligned with human knowledge. Recently, several explanatory interactive machine learning (XIL) methods were developed for this purpose, but each have different motivations and methodological details. In this work, we provide a unification of various XIL methods into a single typology by establishing a common set of basic modules. We discuss benchmarks and other measures for evaluating the overall abilities of XIL methods. With this extensive toolbox, we systematically and quantitatively compare several XIL methods. In our evaluations, all methods are shown to improve machine learning models in terms of accuracy and explainability. However, we found remarkable differences in individual benchmark tasks, which reveal valuable application-relevant aspects for the integration of these benchmarks in the development of future methods.\footnote{{\color{red} Preprint, published article available at: \url{https://www.nature.com/articles/s42256-023-00612-w}}.}
\end{abstract}

\noindent Trust is considered as the ``firm belief in the reliability, truth, or ability of someone or something'' \cite{OxfordLexico_trust}. However, how reliable are ML models and do they in fact base their decisions on correct reasons? 
These questions emerge as ML becomes more present in our daily lives and, importantly,  high-stakes environments, e.g. for disease detection \cite{Obermeyer2019} making it more and more necessary for humans to rely on such machines. However, particularly deep neural networks (DNNs), which are considered state-of-the-art models for many tasks, show an inherent lack of transparency regarding the underlying decision process for their predictions as well as fundamental issues concerning robustness (e.g. slight input perturbations can lead to very different model predictions). Solutions to these issues are considered integral components for trust development in current and future AI systems \cite{holzinger2021_nextAI}.
Particularly, this first issue becomes ever more important for identifying shortcut behavior \cite{Geirhos2020} as the latest trend of DNNs ---large-scale pre-trained models, like GPT-3 and DALL$\cdot$E 2 \cite{GPT3,dalle2}--- employ huge amounts of unfiltered data that contain biases and can lead to negative societal impacts if left unchecked \cite{bender2021stochasticparrots__, Angerschmid2022FairnessAE}.

Consequently, eXplainable AI (XAI) was introduced to address this lack of transparency \cite{belinkov2019analysis,atanasova-etal-2020-diagnostic}. Via such \textit{explainer} methods proposed by XAI research, recent works have revealed that DNNs can show Clever-Hans behavior ---making use of confounders--- due to spurious correlations in the data \cite{Lapuschkin2019CleverHans}. 
However, only making such models explainable can be insufficient for properly building trust as well as for the overall deployability of a model as it does not offer the possibility to revise incorrect and hazardous behavior. For this reason, the eXplanatory Interactive machine Learning (XIL) framework \cite{Teso2019_CAIPI} was proposed in order to promote a more fruitful approach to communication between humans and machines, allowing for a more complementary approach. 

Specifically, in XIL, a model makes a prediction, presents its corresponding explanation to the user, and they respond by providing corrective feedback, if necessary, on the prediction and explanation. 
It has been shown that XIL can improve performance and explanations, i.e. help overcome Clever-Hans behavior and improve the generalization to unseen data \cite{Schramowski2020_Plantphenotyping}.  
Moreover, interaction through explanations is considered a natural form of bidirectional communication between human experts, making XIL methods effective protocols to open black boxes.
In this way, XIL methods may fill the trust gap between ML systems and human users \cite{popordanoska2020}. 

Unfortunately, existing XIL methods \cite{Ross2017_RRR,Schramowski2020_Plantphenotyping,Shao2021_RBR,Rieger2019_CDEP,Selvaraju2019_HINT,Teso2019_CAIPI} were developed independently and often with slightly different motivations. In these works, evaluations of the effectiveness of a method often reverted to qualitative explanation evaluations and test accuracy on separate known confounded data sets. 
However, these evaluation measurements do not unveil essential methodological characteristics that are particularly important for the practical use case. 
Furthermore, currently, no study exists that covers a comprehensive comparison of relevant XIL methods. 
Therefore, in this work, we provide a typology for XIL and propose that existing methods can in fact be summarized via a common underlying terminology. Hereby a method's individual differences correspond to specific instantiations of the basic modules. 
We additionally propose an extensive set of evaluation criteria, consisting of novel measures and tasks, for extensively benchmarking current and future XIL methods based on our typology. This includes the robustness of a method to faulty user feedback and its efficiency in terms of the number of required interactions. Thus, in this work, we provide for the first time an extensive study of six recent XIL methods based on these various criteria.

In summary, our main contributions are (1) unifying existing XIL methods into our typology with a single common terminology, (2) extending the typology by introducing novel measures and tasks to benchmark XIL approaches, (3) evaluating existing methods based on these various criteria that are of great relevance for real-world applicability, and (4) identifying yet unresolved issues to motivate future research.
\begin{algorithm}[t]
    \begin{algorithmic}[1]
            \State $f \gets \textsc{Fit}(\calA)$
            \REPEAT
                \STATE $X \gets \Select(f, \calN)$
                \STATE $\hat{y} \gets f(X)$
                \STATE $\hat{E} \gets \Explain(f, X, \hat{y})$
                \STATE Present $X$, $\hat{y}$, and $\hat{E}$ to the user
                \STATE $\overline{y}, \overline{C} \gets \Obtain(X, \hat{y}, \hat{E})$
                %\STATE $\calA \gets \calA \cup \{(x, \overline{y})\} \cup \{(\overline{x}_i, \overline{y})\}_{i=1}^c$
                \STATE $\calA \gets \calA \cup \{(X, \overline{y}, \overline{C})\}$
                \STATE $f \gets \Revise(\calA) $
                \STATE $\calN \gets \calN \setminus \{X\}$ \label{eq:updatesets}
            \UNTIL{budget $T$ is exhausted or $f$ is good enough}
            \State \textbf{return} $f$
 %       \EndProcedure
    \end{algorithmic}
    \caption{XIL takes as input sets of annotated examples $\calA$ and non-annotated examples $\calN$, and iteration budget~$T$}
    \label{alg:typology}
\end{algorithm}

\section*{Explanatory
Interactive Machine Learning}
To examine eXplanatory Interactive machine Learning (XIL), in the following, we present the first typology for XIL (Fig.~\ref{alg:typo_figure}) based on Algorithm~\ref{alg:typology}. We describe its modules in detail and use them as a foundation for our evaluations. Moreover, we use our typology to examine present XIL methods and thereby investigate the different modules to uncover limitations and motivate avenues for future work.

\subsection*{A Unified XIL Typology}

In general, XIL combines explanation methods (XAI) with user supervision (active learning) on the model's explanations to revise the model's learning process interactively. 
Notably, XIL can be considered a subfield of AI methods that leverage explanations into the learning process (e.g. see \cite{teso22leveraging} for a comprehensive overview of methods that leverage explanations in interactive machine learning)
The conceptual function can be described as follows: XAI focuses on generating explanations from the model, whereas XIL aims to reverse the flow and inserts user feedback on those explanations back into the model. The goal is to establish trust in the model's predictions not only by revealing false and potentially harmful behavior of a model's reasoning process via the model's explanations but also to give the user the possibility to correct this behavior via corrections on these explanations.

\begin{figure}[t]
    % \centering
    % \begin{minipage}{0.48\textwidth}
    % %\begin{algorithm}[b]
    %     \begin{algorithmic}[1]
    %     \setstretch{1.6}
    %         \State $f \gets \textsc{Fit}(\calA)$
    %         \REPEAT
    %             \STATE $X \gets \Select(f, \calN)$
    %             \STATE $\hat{y} \gets f(X)$
    %             \STATE $\hat{E} \gets \Explain(f, X, \hat{y})$
    %             \STATE Present $X$, $\hat{y}$, and $\hat{E}$ to the user
    %             \STATE $\overline{y}, \overline{C} \gets \Obtain(X, \hat{y}, \hat{E})$
    %             %\STATE $\calA \gets \calA \cup \{(x, \overline{y})\} \cup \{(\overline{x}_i, \overline{y})\}_{i=1}^c$
    %             \STATE $\calA \gets \calA \cup \{(X, \overline{y}, \overline{C})\}$
    %             \STATE $f \gets \Revise(\calA) $
    %             \STATE $\calN \gets \calN \setminus \{X\}$ \label{eq:updatesets}
    %         \UNTIL{budget $T$ is exhausted or $f$ is good enough}
    %         \State \textbf{return} $f$
    %  %       \EndProcedure
    %     \end{algorithmic}
    %     % \captionof{algorithm}{XIL takes as input sets of annotated examples $\calA$ and non-annotated examples $\calN$, and iteration budget~$T$}
    %     % \label{alg:typology}
    % %\end{algorithm}
    % \end{minipage}
    % \begin{minipage}{0.5\textwidth}
    \centering
    \includegraphics[width=0.3\textwidth]{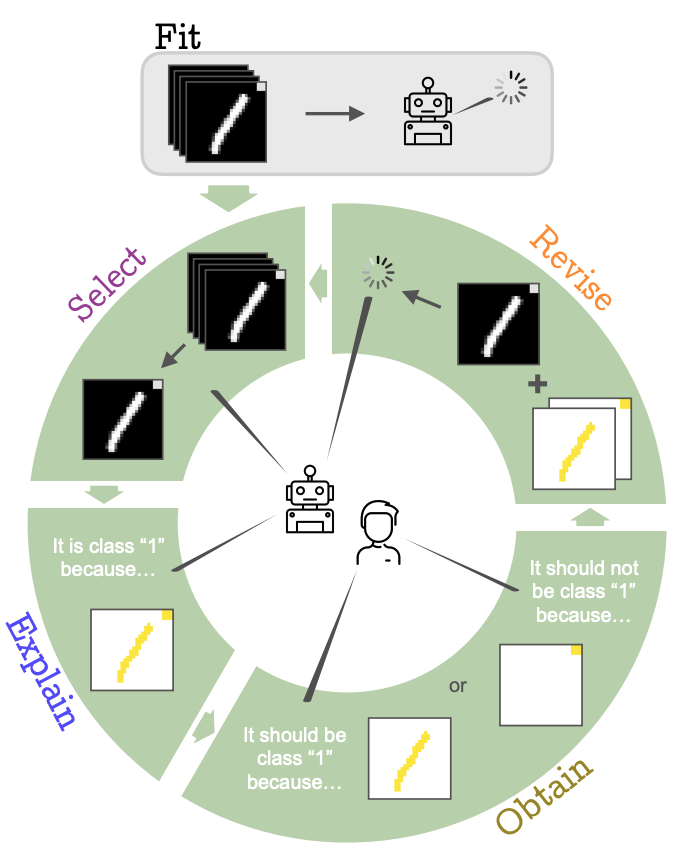}
    \caption{The XIL Typology. Flowchart visualizing Algorithm~\ref{alg:typology}. \Select \xspace describes how samples $X$ from $\calN$ are selected in XIL. \Explain \xspace depicts how the model provides insights into its reasoning process to the teacher. With \Obtain, the teacher, in turn, observes whether the learner's prediction is right or wrong, especially if it is based on the right or wrong reason, and returns corrective feedback, if necessary. The (explanatory) corrections obtained are redirected back into the model's learning process with the \Revise \xspace module to correct the model behavior according to the user.}
    \label{alg:typo_figure}
\end{figure}

Algorithm \ref{alg:typology} describes the XIL setting in pseudo-code. It uses a set of annotated examples $\calA$, a set of non-annotated examples $\calN$, and an iteration budget $T$. The annotation comprises both the classification label $y$ and explanation $E$, i.e. a non-annotated example is missing one or both. In general, the procedure is illustrated in Fig.~\ref{alg:typo_figure} and can be compared to a teacher-learner setting. 
Active learning is a learning protocol in which the model sequentially presents non-annotated examples (\Select) from a data pool to an oracle (e.g. human annotator) that labels these instances (\Obtain).
Accordingly, this setting allows the user to influence the learning process actively (\Revise). 
Although the active learning setting enables simplistic interaction between the model and a user, it does not promote trust if explanations do not accompany predictions \cite{Teso2019_CAIPI}. The lack of explanations in active learning, however, makes it difficult for the user to comprehend the model's decision process and provide corrections. Therefore, the XIL typology extends the learning pipeline with XAI (\Explain). Consequently, the explanations and potential user corrections are processed simultaneously with the annotated labels. The necessary modules of this interactive learning loop (Fig.~\ref{alg:typo_figure}) are each described in detail below.

\subsubsection*{\textbf{Selection} (\Select)} 
\Select \xspace describes how samples $X$ are selected from a set of non-annotated examples $\calN$. These examples are used for the model to perform a predictive task, e.g. predict a class label $y$, with which the user, in turn, has to interact. The selection can be carried out in different ways: manually, randomly, or with a specific strategy. One strategy in this regard is to find influential examples, e.g. via a model's certainty in a prediction. This can also enable selecting only a subset of examples to apply XIL on. Hence, \Select \xspace also describes how many examples need to be selected to revise a model through explanatory interactions.

\subsubsection*{\textbf{Explaining} (\Explain)}
In comparison to active learning, XIL approaches consider standard input-output pairs, e.g. $(X, \hat{y})$, insufficient to (i) understand the underlying decision process of a model and (ii) provide necessary feedback solely on the predicted labels, denoted as $\overline{y}$. Such feedback, $\overline{y}$, can only correct the model if the model's initial prediction, $\hat{y}$, is incorrect, i.e. \textit{wrong answer}. Due to, e.g., shortcut learning \cite{Geirhos2020}, deeper insights into a model are required. Hence, in XIL, the model also provides explanations that help the user inspect the reasoning behind a prediction. This, in turn, enables a user to check if the decision is based on \textit{right} or \textit{wrong reasons}. Therefore, \Explain \xspace is an essential element of a XIL method to revise a model.

In our proposed typology, the learner $f$ (e.g. a CNN) predicts $\hat{y}$ for an input $X$. Additionally, the learner explains its prediction to the teacher (e.g. user) via an explainer (e.g. LIME) and provides an explanation $\hat{E}$. 
In this way, \Explain \xspace depicts how the model provides insights into its reasoning process to the teacher. 

There are various ways to provide an explanation.
Common explanation methods in works of XIL provide attribution maps that highlight important features in the input space, such as input gradients (IG \cite{hechtlinger2016interpretation}), gradient-weighted class activation maps (GradCAM \cite{Selvaraju2016_GradCAM}), and local interpretable model-agnostic explanations (LIME \cite{Ribeiro2016_LIME}).

\Explain \xspace also describes the capability of a XIL method to facilitate the use of various explainer methods, i.e. whether a XIL method depends on a specific explainer method. Whereas some XIL methods can handle arbitrary explainer methods (e.g. CE), it is the defining component for other XIL methods and thus constrains other components of the method as well (e.g. feedback types). 

Analogous to the view on the explainers, the model flexibility describes the capability of a XIL method to facilitate the use of different model types for \Explain \xspace. Depending on the used model, only specific XAI methods can be applied, e.g. whereas LIME can be applied to any ML model, IG can only be applied to differentiable ones (e.g. NNs), and GradCAM only to CNNs. In turn, this means that a XIL method can be model-specific or model-agnostic. However, the model specificity is linked to the explainer specificity as an explainer may be only available for certain model types.

\subsubsection*{\textbf{Obtain Feedback} (\Obtain)}
Not only does the model have to explain its decision, but also the users have to provide explanatory feedback to the model. This feedback has to be processed in such a way that the model can cope with it. As a result, the model can generate corrections based on user feedback to revise the model.
The correction $\overline{C}$ depends on the specific XIL method and model type. Specifically, the user's feedback $\overline{C}$, wrt. the explanation $\hat{E}$, has to potentially be converted to an input space that the model can process.
For instance, in the case of counterexamples, the user feedback $\overline{E}$ is on the same level as the explanation, e.g. an attribution map. However, correction $\overline{C}$ depicts one or multiple counterexamples, such that $\overline{E}$ must be converted.

In our setup, the teacher gives feedback based on the model's input $X$, prediction $\hat{y}$, and explanation $\hat{E}$. Specifically, within \Obtain, the teacher produces a corresponding explanation, $\overline{E}$, which, however, is transformed into a feedback representation, $\overline{C}$, that corresponds to a representation that can be fed back to the learner. This enables the teacher to observe whether the learner's prediction is right or wrong, but more importantly also to check if the prediction is based on the right or wrong reason.

Moreover, \Obtain \xspace determines which feedback types a XIL method can handle. The standard way to provide feedback, partly restricted by using attribution maps in XAI, is to highlight important (right) and/or unimportant (wrong) features in the input. 
Although, other types of feedback are also possible, e.g. in the form of semantic description, e.g. ``Never base the decision on the shape of object X''\cite{stammer2021}.

\subsubsection*{\textbf{Model Revision} (\Revise)}
Once the corrections are obtained, they must be redirected back into the model's learning process. Depending on the feedback type and the user's knowledge about what is right or wrong, there are two aspects to consider to revise a model. 

The first aspect is how to reinforce user feedback. As indicated in \Obtain \xspace, the \Revise \xspace strategy depends on the feedback obtained from the user. On the one hand, the user can penalize wrong explanations, i.e. remove confounding factors but not necessarily guide the model towards the right reason. On the other hand, the user can reward the right explanations. Intuitively, it is harder to know the right reason than the wrong reason e.g. on average, the subspace of relevant image regions is much smaller than the space of irrelevant ones). Additionally, rewarding does not necessarily ensure avoiding a confounder's influence. In general, there therefore seems to be an imbalance between knowing what is right and wrong, which needs to be considered. 

The second aspect is how to update the model, i.e. incorporate the feedback. One common approach is to augment the loss function and backpropagate the feedback information through the loss objective. The other is to augment the dataset with (counter)examples and remove the confounder influence through a diminished presence in the training data.

After the teacher gives feedback to the learner, the corrections are fed back to the learner to revise it. To do so, set $\calA$ is extended by the processed user annotations, i.e. the prediction $\overline{y}$ and the correction $\overline{C}$ for the respective input $X$. The optimization objective can now incorporate the user feedback to extend the purely data-driven approach and thereby revise (fit) the model $f$. Lastly, \calN~is updated, i.e. the annotated instances $X$ are removed from \calN.

\subsection*{No Free Lunch in XIL}

We hypothesize that there is no single best XIL method. Changing a module has costs such that a modification may increase the performance in one criterion but at the expense of another. Hence, we investigate the different modules with various experiments to verify our hypothesis. 
Moreover, we showcase our typology and the corresponding evaluation criteria by benchmarking, for the first time, existing XIL methods and their modules. By providing a comprehensive evaluation of these methods, we also reveal yet undiscovered limitations to encourage future research.

To this end, we investigate the following questions: 
\textbf{(Q1)} How well do the existing methods revise a model? 
\textbf{(Q2)} Are they robust to feedback quality variations? 
\textbf{(Q3)} Does the revision still work with a reduced number of interactions?
\textbf{(Q4)} Can XIL revise a strongly confounded model?

\paragraph{XIL Methods, Measures and Benchmarks}
We focus our evaluations on computer vision datasets, where confounders are well-known and an active area of research \cite{ZhongConfounder}. In the relevant datasets, a confounder is a visual region in an image (e.g. colored corner) correlating with the image class but is not a causal factor for determining the true class of the image. The confounder fools the model and constitutes a shortcut learning rule \cite{Geirhos2020}. In the standard setup, we train an ML model on a confounded train set and run tests on the non-confounded test set. Our goal is to guide the model to ignore the confounder. To account for different facets of XIL, we chose two benchmark datasets, Decoy(F)MNIST, and one scientific dataset, ISIC19. For these datasets, a confounder is visual (in the sense that they are spatially separable) to provide a controlled environment for evaluation. 
\begin{table}[t]
    \centering
    \def\arraystretch{1}\tabcolsep=2.pt
    \begin{tabular}{l|cccccc}
     Module & RRR & RRR-G & RBR & CDEP & HINT & CE \\ \hline
     Select & \multicolumn{6}{c}{random} \\
     Explain & IG & GradCAM & IF & CD & GradCAM & LIME \\
     Obtain & \multicolumn{6}{c}{attribution mask} \\
     Revise & penalty & penalty & penalty & penalty & reward & dataset\\
     \end{tabular}
     \caption{Overview of the XIL methods setup in our experiments: RRR \cite{Ross2017_RRR}, RRR-G \cite{Schramowski2020_Plantphenotyping}, RBR \cite{Shao2021_RBR}, CDEP \cite{Rieger2019_CDEP}, HINT \cite{Selvaraju2019_HINT} and CE \cite{Teso2019_CAIPI}.}
    \label{tab:xil_summary}
\end{table} 

In the following, we evaluate the XIL methods: (i) RRR (Right for the Right Reason \cite{Ross2017_RRR}), (ii) CDEP (Contextual Decomposition Explanation Penalization \cite{Rieger2019_CDEP}), (iii) HINT (Human Importance-aware Network Tuning \cite{Selvaraju2019_HINT}), and (iv) CE (CounterExamples \cite{Teso2019_CAIPI}) and analyze the influence of different explainers on the same method, namely IF referenced to as (v) RBR (Right for the Better Reasons \cite{Shao2021_RBR}) and GradCAM in the following called (vi) RRR-G (Right for the Right Reason Gradcam \cite{Schramowski2020_Plantphenotyping}). We summarize all investigated methods with their respective implementation of each component in Tab.~\ref{tab:xil_summary}.
We set up novel measures and benchmarks in the Methods section to provide detailed insights into the versatile facets of a XIL method. Besides standard measures like accuracy, we provide a new measure, \wrnospace, to investigate a model's focus on the wrong reason(s). Furthermore, we provide extensive benchmarks such as the feedback robustness, the interaction efficiency, and a \textit{Switch XIL on} experiment. These benchmarks help examine various essential aspects of a XIL method beyond simple accuracy scores.
More details on these and the experimental protocol can be found in Methods.
\begin{table}[t]
    \begin{subtable}{\textwidth}
    \centering
    \begin{tabular}{cl|l}
        \multicolumn{1}{c}{} & \multicolumn{2}{c}{DecoyMNIST} \\
        XIL & train & test \\ \hline 
        w/o decoy & $99.8 {\pm0.1}$ & $98.8 {\pm0.1}$ \\ \hline
        Vanilla & $99.9 {\pm0.0}$ & $78.9 {\pm1.1}$ \\ \hline
        RRR & $99.9 {\pm0.1}$ & $98.8 {\pm0.1}$ \\
        RRR-G & $99.7 {\pm0.2}$ & $97.4 {\pm0.7}$ \\
        RBR & $\mathbf{100}  {\pm0.0}$ & $\mathbf{99.1} {\pm0.1}$  \\
        CDEP & $99.3 {\pm0.0}$ & $97.1 {\pm0.7}$ \\
        HINT & $97.6 {\pm0.3}$ & $96.6 {\pm0.4}$ \\
        CE & $99.9 {\pm0.0}$ & $98.9 {\pm0.2}$ \\ 
    \end{tabular}
    \quad
    \begin{tabular}{l|l}
        \multicolumn{2}{c}{DecoyFMNIST} \\
        train & test \\ \hline
        $98.7 {\pm0.3}$ & $89.1 {\pm0.5}$ \\ \hline
        $99.5 {\pm0.2}$ & $58.3{\pm2.5}$\\ \hline
        $98.7 {\pm0.3}$ & $\mathbf{89.4} {\pm0.4}$ \\ 
        $90.2 {\pm1.6}$ & $78.6 {\pm4.0}$ \\ 
        $96.6 {\pm2.3}$ & $87.6 {\pm0.8}$ \\ 
        $89.8 {\pm2.7}$ & $76.7 {\pm3.5}$  \\ 
        $99.0 {\pm0.9}$ & $58.2 {\pm2.3}$ \\ 
        $\mathbf{99.1} {\pm0.2}$ & $87.7 {\pm0.8}$ \\
    \end{tabular}
    \quad
    \begin{tabular}{l|l}
        \multicolumn{2}{c}{ISIC19 } \\
        train & test\\
        \hline
        -- & -- \\
        \hline
        $100{\pm0.0}$ & $88.4{\pm0.5}$ \\
        \hline
        $\mathbf{100}{\pm0.0}$ & $88.1{\pm0.4}$ \\
        $\mathbf{100}{\pm0.0}$ & $\mathbf{88.4}{\pm0.5}$\\
        $92.6{\pm5.3}$ & $80.3{\pm5.6}$ \\
        $\mathbf{100}{\pm0.0}$ & $87.9{\pm0.5}$ \\
        $\mathbf{100}{\pm0.0}$ & $87.7{\pm0.5}$ \\
        $\mathbf{100}{\pm0.0}$ & $87.5{\pm0.5}$ \\
    \end{tabular}
    % \caption{Mean accuracy scores [\%]. First row shows performance on dataset without decoy squares (not available for ISIC19). Next row shows the Vanilla model (no XIL) gets fooled, indicated by low test accuracy. Except for HINT on FMNIST, all methods recover test accuracy. On ISIC19, no accuracy improvement can be observed. Best values are bold; cross-validated on 5 runs with std.}
    \caption{Accuracy}
    \label{tab:exp_performance}
% \end{table}
    \end{subtable}
    \vspace{0.3cm}
    
% \begin{table}[t]
%     \centering
    \begin{subtable}{\textwidth}
    \centering
    \def\arraystretch{1}\tabcolsep=2.pt
    \begin{tabular}{cl|l|l}
        \multicolumn{1}{c}{} & \multicolumn{3}{c}{DecoyMNIST} \\
        XIL & IG & GradCAM & LIME \\ \hline 
        Vanilla & $23.1 \pm3.8$ & $38.7 \pm4.6$ & $59.8\pm2.0$ \\ \hline
        RRR & $\mathbf{0.0} \pm0.0$ & $13.3 \pm2.0$& $\mathbf{32.1}\pm0.4$\\ 
        RRR-G & $11.9 \pm2.1$& $\mathbf{1.5} \pm0.8$ & $33.3\pm2.8$\\ 
        RBR & $2.0 \pm1.3$ & $15.2 \pm3.8$ & $37.7\pm 3.0$\\
        CDEP & $15.0 \pm1.5$& $27.8 \pm3.8$& $37.9\pm3.7$\\ 
        HINT & $11.9 \pm3.1$ & $46.8 \pm1.1$& $53.8\pm2.0$\\ 
        CE & $7.3 \pm1.4$ & $14.7 \pm2.9$ & $36.9\pm0.6$\\
    \end{tabular}
    % \quad
    \hspace{2pt}
    \begin{tabular}{l|l|l}
        \multicolumn{3}{c}{DecoyFMNIST} \\
        IG & GradCAM & LIME \\ \hline 
        $25.0 \pm1.9$ & $34.8 \pm1.4$& $57.6\pm0.8$ \\  \hline
        $\mathbf{0.0} \pm0.0$ & $24.2  \pm4.1$& $\mathbf{27.4}\pm0.7$\\
        $2.1 \pm0.4$& $\mathbf{4.6} \pm0.9$ & $38.1\pm4.5$\\ 
        $5.97 \pm1.4$& $16.0 \pm4.8$ & $34.9\pm1.4$\\ 
        $15.9 \pm4.5$& $39.1 \pm1.7$ & $40.2\pm6.5$\\ 
        $29.4 \pm3.3$& $27.8 \pm2.9$& $51.4\pm3.5$ \\ 
        $8.1 \pm0.4$& $24.4 \pm0.9$ & $31.1\pm0.6$\\
    \end{tabular}
    % \quad
    \hspace{2pt}
    \begin{tabular}{l|l|l}
        \multicolumn{3}{c}{ISIC19} \\
        IG & GradCAM & LIME \\ \hline 
        $33.2\pm0.2$ & $35.2\pm0.8$ & $63.6\pm0.7$ \\ \hline 
        $16.6\pm8.7$ & $27.4\pm4.3$ & $58.9 \pm 1.4$ \\ %\textbf{34.9}\\ 
        $\mathbf{11.9} \pm0.9$ & $\mathbf{0.9}\pm0.1$ & $\mathbf{34.7}\pm0.8$\\ 
        $17.2\pm1.4$ & $28.5\pm22.7$ & $58.0\pm0.6$ \\
        $25.5\pm0.2$ & $5.4\pm0.2$ & $67.4\pm0.0$\\ 
        $31.1\pm0.1$ & $22.0\pm0.2$  & $60.9\pm0.0$ \\ 
        $32.7\pm 0.0$ & $36.6\pm0.1$ & $61.5\pm0.9$\\
    \end{tabular}
    \caption{\wrmeasure}
    \label{tab:exp_wr}
    \end{subtable}
    \caption{Mean accuracy scores and \wrmeasure scores each in [\%]; best values are bold; cross-validated on 5 runs with standard deviation. (a) First row shows performance on dataset without decoy squares (not available for ISIC19). Next row shows the Vanilla model (no XIL) gets fooled, indicated by low test accuracy. Except for HINT on FMNIST, all methods recover test accuracy. On ISIC19, no accuracy improvement can be observed; higher is better. (b) XIL reduces \wrmeasure scores for all methods on all datasets, even for ISIC19; lower is better.}
    \label{tab:exp_performance_overall}
\end{table}

\paragraph{(Q1) Accuracy Revision} In order to investigate the general ability of a XIL method to revise a model (\Revise), we evaluate the accuracy score on each test set (Tab.~\ref{tab:exp_performance}) of the datasets DecoyMNIST, DecoyFMNIST and ISIC19. To give an impression of the confounder impact in each dataset, we provide a baseline by evaluating each model on the dataset without decoy squares. This is not available for ISIC19 as the confounders are not artificially added. The Vanilla model represents the performance of a model without revision via XIL. The confounder leads the models to be fooled, causing low accuracy scores compared to the baseline without the decoy. 

In contrast, training via the examined XIL methods generally helps a model overcome the confounder, as the baseline test accuracy is recovered. RBR performs best on DecoyMNIST and RRR on DecoyFMNIST. For DecoyFMNIST, HINT achieves a low accuracy score on par with the fooled Vanilla model, indicating that it, here, cannot correct the Clever-Hans behavior. We assume that its reward strategy does not suffice to overcome the confounder and, in turn, for XIL to function properly. For the ISIC19 dataset, no XIL method helps a model improve the accuracy on the test set. Therefore, we cannot answer (Q1) affirmatively for ISIC19 purely based on the accuracy, thus motivating the need for further measures beyond accuracy.

However, summarized, our experiments answer (Q1), i.e. the XIL methods have the general ability to revise a model but may have difficulties with increasing data complexity.

\begin{figure}[t]
  \centering
  \includegraphics[width=0.49\textwidth]{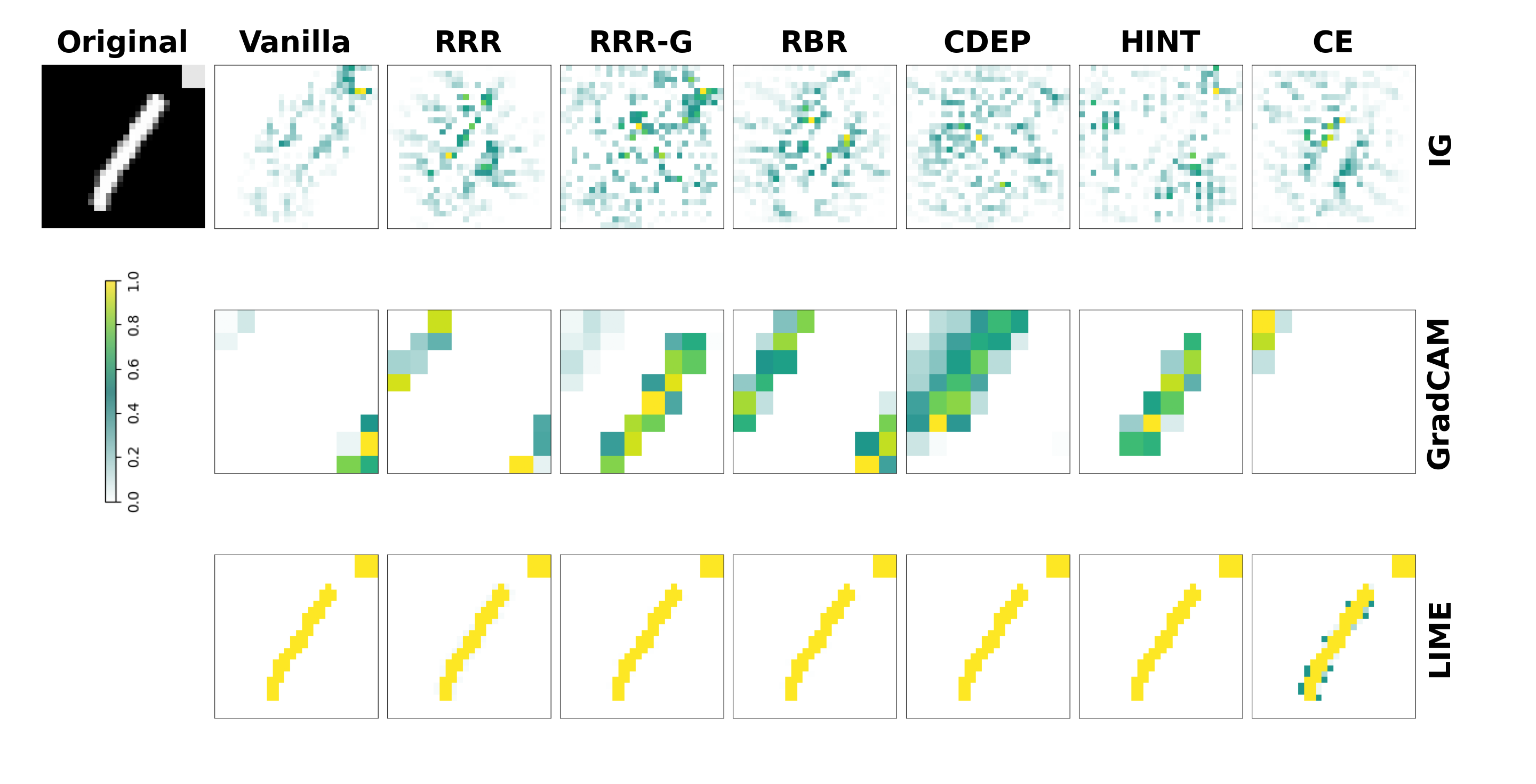}
  \includegraphics[width=0.49\textwidth]{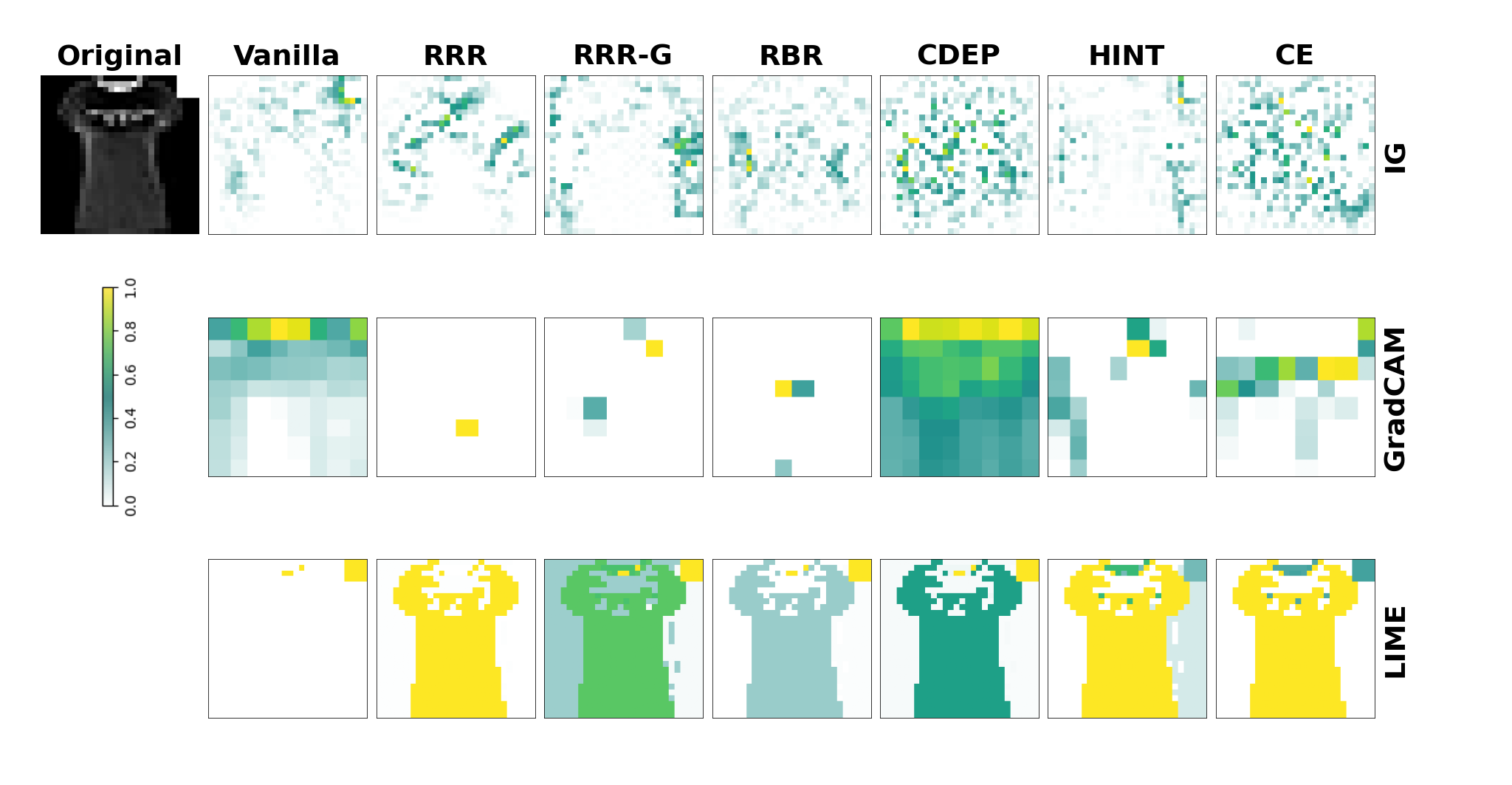}
  \caption{Qualitative inspection of explanations. The first column on each side shows the original image, the second column shows the Vanilla model (no XIL) attribution maps, and the remaining columns show the attribution maps of a model with each XIL method. Each row represents an explanation method to visualize the model prediction. The color bar indicates the activation of attribution maps (yellow (1) represents max activation and white (0) min activation). On the left (right) results for DecoyMNIST (DecoyFMNIST).}
  \label{fig:MNISTheatmap}
\end{figure}

\paragraph{(Q1) Wrong Reason Revision}
For the ability to revise wrong reasons, we conduct quantitative (Tab.~\ref{tab:exp_wr} and qualitative (Fig.~\ref{fig:MNISTheatmap}) experiments to inspect \Explain.

On one hand, we have the quantitative \wrmeasure score. It measures the activation in the confounder area and hence automates the visual inspection of explanations. The Vanilla model (without XIL) achieves high \wrmeasure scores, i.e., high activation in the confounder region. Again, the XIL methods help a model lower the \wrmeasure score, reducing the confounder impact. The table further shows that the XIL methods overfit on the internally-used explainer in terms of reducing its attention to the confounding region (cf. RRR with IG explanations or RRR-G with GradCAM explanations). This is expected, as a XIL method exactly optimizes for the used explanation method. Interestingly, a XIL method also reduces the \wrmeasure score for other not internally-used explainers (cf. RRR with GradCAM explanations or RRR-G with IG explanations). Consequently, XIL's impact is beyond its internally-used explainer and not only restricted to it. However, LIME attribution maps are always highly activated, albeit reduced, as it is never internally used as an explainer. 

Furthermore, we can see that CDEP and HINT do not remarkably reduce the \wrmeasure score compared to the baseline. As HINT works with rewarding instead of penalizing and is thus not explicitly trained to avoid confounders, we do not necessarily expect it to score low \wrmeasure values. CDEP also does not achieve low \wrmeasure values and does not overcome the confounder, although using penalty.
We previously found that XIL could not overcome the test performance of the fooled Vanilla model on the ISIC19 dataset. However, the \wrmeasure score surprisingly indicates a clear reduction. This indicates that although XIL might not revise a model in terms of accuracy, it can indeed improve the explanations (lowers the \wrmeasure score), proving its function and usefulness. Notably, these findings showcase the importance of additional quantitative measures, such as \wrnospace, for evaluating XIL methods.

Apart from this, we manually inspected 200 randomly generated attribution maps for each method-explainer combination. We exemplify the findings for a representative example on each benchmark dataset.
Fig.~\ref{fig:MNISTheatmap} shows explanation attribution maps for Decoy(F)MNIST on the left (right). A high activation in or around the top right corner indicates Clever-Hans behavior of a model to the confounder. For the Vanilla model, each explanation method highlights the confounder region, except for the GradCAM explanation on MNIST (\cf Discussion). The top row shows activation attribution maps generated with the IG explainer. Here, we can see the previously-found overfitting once again. For example, the RRR-revised model shows no activation in the confounding right top corner while RRR-G still has high activation around the corner. Consequently, XIL functions only reliably on the internally-used explainer. The qualitative findings here confirm the quantitative findings of the \wrmeasure score and demonstrate that it is a suitable method to evaluate the performance of XIL methods.

Moreover, in our evaluation, we also find that penalizing wrong reasons does not enforce predictions based on the right reasons (cf. RRR-G with IG attribution maps on MNIST).
Second, attribution maps generated via GradCAM require upsampling and hence prevent a clear and precise interpretation. Although the right reason is sometimes highlighted, there remains some uncertainty in its precision.
Third, the provided explanation methods visually contradict each other. The RRR column, for instance, indicates that XIL, and XAI in general, must be handled with care: the performance values may show overcoming a confounder while the visual explanations (attribution maps) indicate otherwise.

Overall, our evaluation gives more insight into \Explain \xspace of XIL and extends the previous findings for (Q1). Although it may not become entirely apparent that the considered XIL methods remove all Clever-Hans behavior when only considering accuracy, we observed that the methods and, in this way, XIL in general do in fact improve a model's explanations and can therefore effectively be used to revise a model.

We note here that we found possibly additional confounding factors in the ISIC19 dataset and hence focus further evaluations on the remaining two datasets. 

\paragraph{(Q2) Robustness to feedback quality variations}
As previous research focused only on providing correct (ground-truth) feedback, we additionally provide insights into the feedback quality and the robustness of a XIL method towards quality changes. In this experiment, the objective is to gather more knowledge about \Obtain.

\begin{table}[t]
    \centering
    \def\arraystretch{1}\tabcolsep=4.pt    
    \begin{tabular}[t]{c|c|c|c|c|c|c|c|c|c|c|c|c}
        \multicolumn{1}{c}{} & \multicolumn{6}{c|}{DecoyMNIST} & \multicolumn{6}{c}{DecoyFMNIST}\\ 
        feedback & RRR & RRR-G & RBR & CDEP & HINT & CE & RRR & RRR-G & RBR & CDEP & HINT & CE \\ \hline
        arbitrary -- & \cellcolor{green!20}+$3.3$ & \cellcolor{green!20}--$4.2$ & \cellcolor{red!40}--$22.1$ & \cellcolor{red!40}+$17.8$ & \cellcolor{green!20}+$4.1$ & \cellcolor{green!40}+$0.3$ & 
        \cellcolor{green!40}+$1.4$ & \cellcolor{red!20}+$7.9$& \cellcolor{red!40}--$37.3$ & \cellcolor{red!40}+$12.2$& \cellcolor{green!20}--$3.2$& \cellcolor{green!40}--$1.2$ \\
        incomplete $\uparrow$ & \cellcolor{green!40}+$19.6$ & \cellcolor{green!40}+$9.5$ & \cellcolor{green!40}+$17.2$ & \cellcolor{green!40}+$17.9$ & \cellcolor{green!40}+$17.7$ & \cellcolor{green!20}+$6.7$ & \cellcolor{green!40}+$24.2$ &\cellcolor{green!40}+$12.4$ & \cellcolor{green!40}+$16$ & \cellcolor{green!40}+$16.8$& \cellcolor{green!40}+$21$& \cellcolor{green!20}+$3.4$ \\
        correct $\uparrow$ & \cellcolor{green!40}+$19.9$ & \cellcolor{green!40}+$18.5$ & \cellcolor{green!40}+$20.2$ & \cellcolor{green!40}+$18.2$ & \cellcolor{green!40}+$17.7$ & \cellcolor{green!40}+$20$ & \cellcolor{green!40}+$31.1$ & \cellcolor{green!40}+$20.3$ & \cellcolor{green!40}+$29.3$ & \cellcolor{green!40}+$18.4$ & \cellcolor{red!20}--$0.1$ & \cellcolor{green!40}+$29.4$\\
    \end{tabular}
    \caption{Feedback robustness evaluation on Decoy(F)MNIST for arbitrary and incomplete compared to correct feedback masks. The mean difference in test accuracy [\%] compared to the confounded Vanilla model is given. For arbitrary feedback, unchanged is better, and for incomplete and correct feedback, higher is better. Incomplete feedback is on par with correct feedback. The values are cross-validated on 5 runs (cf. Supplement Tab.~5 for standard deviations).}
	\label{tab:feedback}
\end{table} 

Tab.~\ref{tab:feedback} compares the impact of different feedback types to a fooled Vanilla model. The values for correct feedback are taken from Tab.~\ref{tab:exp_performance}. Correct feedback demonstrates how XIL improves the accuracy, i.e. removes the confounder impact, compared to the Vanilla model. Moreover, we can clearly see that incomplete and correct feedback are nearly on par for all methods in improving the test accuracy. This emphasizes XIL's robustness towards user feedback of varying quality and suggests real-world usability of XIL, considering that human user feedback is prone to errors. Note, the performance of CE for incomplete feedback is worse due to the strategy of augmenting the dataset. While all confounded images remain in the data, also the added images to revise the model still contain part of the confounder. This way, the confounder impact is still quite high and thus not as easily removed by adding images where only part of the confounding square is removed. However, our results indicate that this still suffices to achieve limited revision. 

In contrast, for the case of arbitrary feedback, \textit{robustness} expresses that a method is not remarkably changing performance. However, we can see a remarkable increase (decrease) for CDEP (RBR) especially compared to the correct feedback improvement. This suggests that CDEP improves performance no matter what feedback quality. Consequently, we presume CDEP does not pass the sanity check leaving some concerns about its reliability. If it is irrelevant what the user feedback looks like to correct the model, the rationale behind the XIL method is questionable and its usage worrisome for users. For RBR, we presume that arbitrary feedback leads to a collapse of the model's learning process, i.e. random guessing, revealing a lack of robustness. A method sensitive to wrong feedback ---humans are not perfect--- can be assessed as worse than a robust method. 

All in all, however, the considered XIL methods prove general robustness for different feedback quality types, thus answering (Q2) affirmatively and providing evidence for XIL's effectiveness in more practical use cases.
\begin{figure}[t]
	\centering
    \vspace{0.3cm}
	\begin{subfigure}{.7\textwidth}
        \centering
        \includegraphics[width=\linewidth]{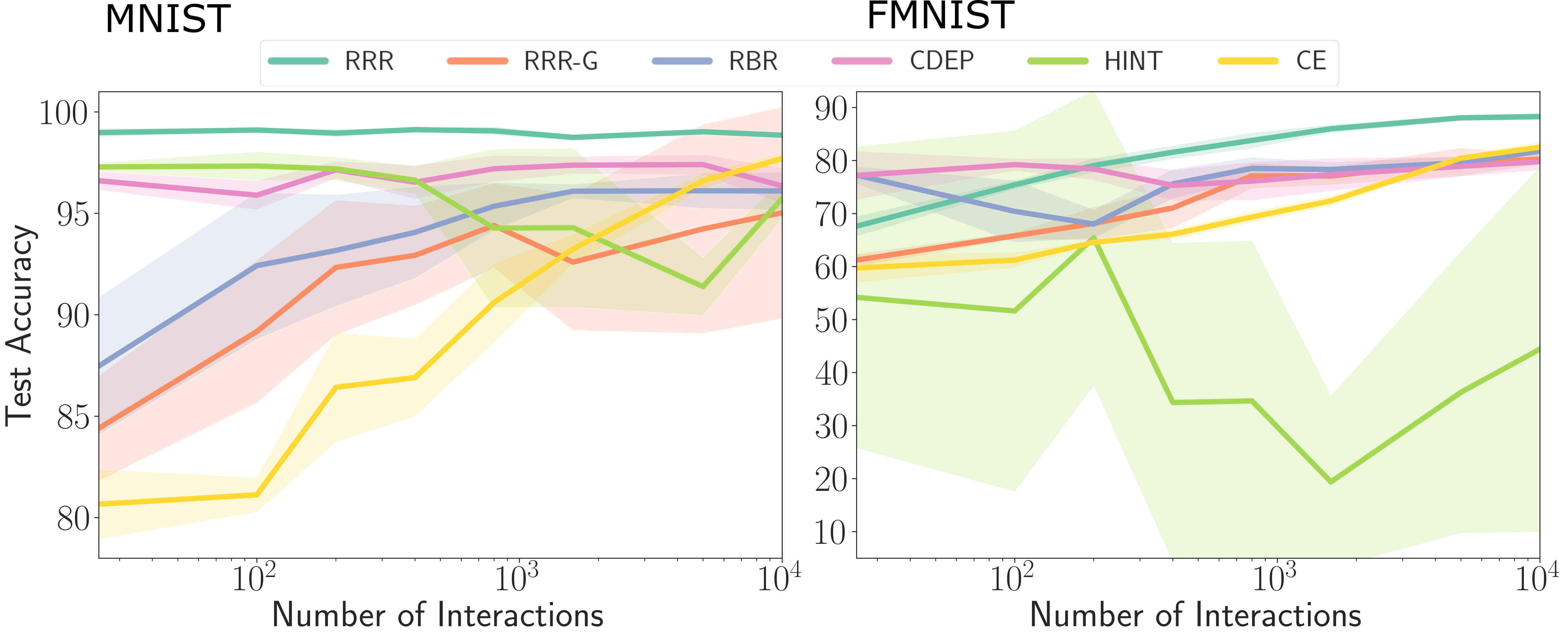}
    	\caption{Interaction Efficiency}
    	\label{fig:inf_eff}
    \end{subfigure}
	\hspace{4pt}
    \begin{subfigure}{.7\textwidth}
	    \centering
        \includegraphics[width=\linewidth]{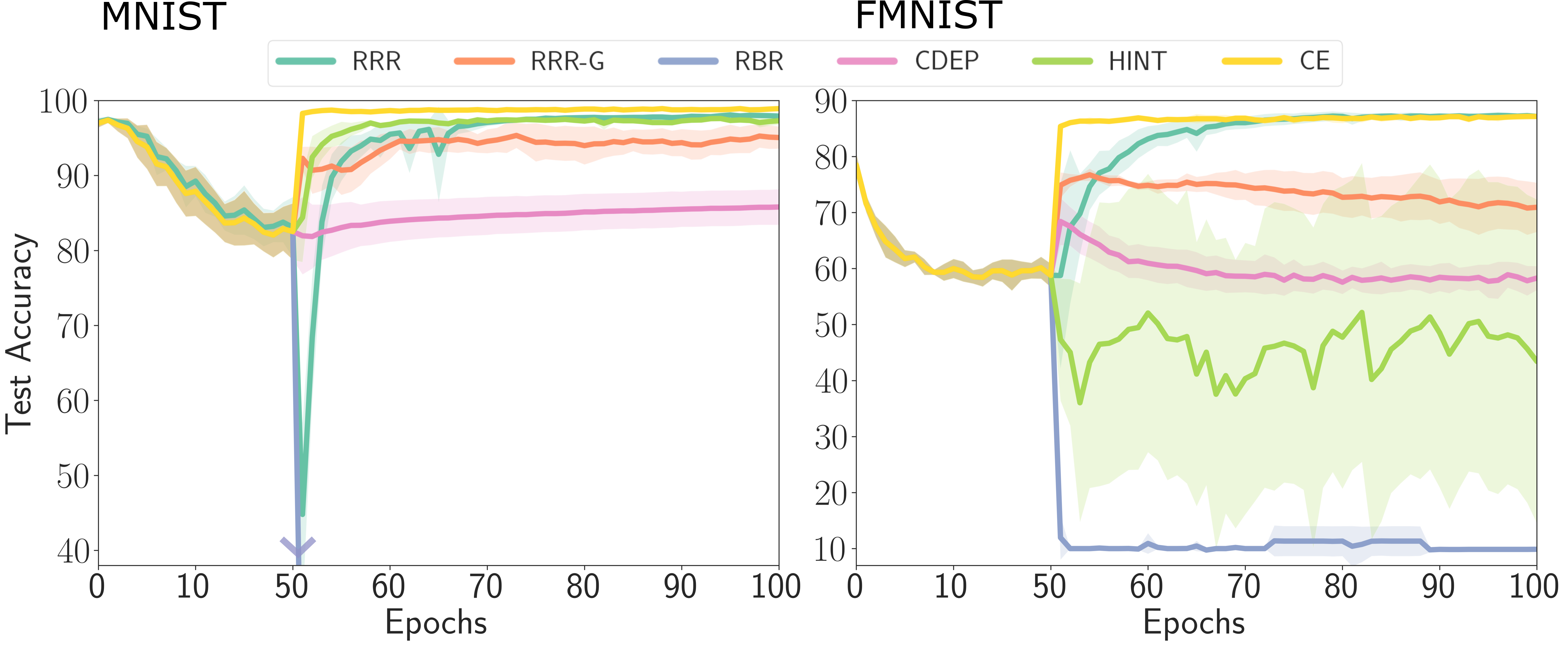}
    	\caption{Switch XIL on}
    	\label{fig:xilon}
    \end{subfigure}
    \caption{Evaluation of (a) interaction efficiency and (b) performance in unconfounding a pre-trained model. Each task is evaluated on the DecoyMNIST (left) and DecoyFMNIST (right) dataset. (a) Test accuracy [\%] with different numbers of used feedback interactions. The more interactions, the better the performance. However, a smaller number of interactions already suffices. (b) Test accuracy [\%] over time after XIL is applied to an already fooled model. All methods, except CDEP and RBR, can recover the test performance and overcome the confounder. Each value represents the mean performance cross-validated on 5 runs; the given confidence intervals represent the standard deviation; (left in c) the arrow indicates the curve drops to random performance.}
\end{figure}

\paragraph{(Q3) Interaction efficiency}
Obtaining precise and correct feedback annotations is costly and time-consuming, making interaction efficiency a crucial property for XIL methods. Therefore, we examine how many explanatory interactions suffice to overcome a known confounder. A method that utilizes annotations more efficiently, i.e., requires fewer interactions to revise a model, is preferable. In the previous experiments, every training image was accompanied by its corresponding feedback mask to correct the confounder. In contrast, now we randomly sample a subset of $k$ annotations before training and evaluate each model with different-sized feedback sets, i.e., number of explanatory interactions. By doing so, we target \Select \xspace as we investigate how the selection affects the model revision.

Fig.~\ref{fig:inf_eff} shows increasing test accuracy for an increasing number of available feedback masks for all XIL methods, i.e. the more feedback available, the better. Moreover, the figure shows that only a tiny fraction of feedback masks is required to revise a model properly. Although there is a remarkable difference between the XIL methods' efficiency, our obtained results illustrate that XIL utilizes feedback efficiently and can already deal with a few feedback annotations. 
Note that the methods achieve different test accuracy with all available feedback, such that they do not all converge at the same level; cf. Tab.~\ref{tab:exp_performance} for test accuracy with full feedback set. Interestingly, RRR, for example, needs only a few dozen interactions to overcome the confounder, while CE requires considerably more interactions. 
Summarized, the examined XIL methods can efficiently utilize user feedback, solving (Q3).

\paragraph{(Q4) Revising a strongly corrupted model}
In order to further evaluate the real-world usability of XIL, we conduct a \textit{Switch XIL on} experiment, where we integrate a XIL method (\textit{switch XIL on}) after a model has solely been trained in a baseline setting (e.g. standard classification) and shows strong Clever-Hans behavior. Fig.~\ref{fig:xilon} shows the test performance of a model during training. First, the model gets fooled by the decoy squares. After 50 epochs, the XIL augmentation is switched on (i.e., either the loss or dataset is augmented with XIL). As we can see, all methods, except CDEP and RBR, can recover the test accuracy and overcome the confounder. RRR shows a striking curve with the RR loss sharply increasing, and hence the accuracy drops before it sharply increases again. Most likely, it requires more hyperparameter tuning to avoid this leap. For RBR, we assume the same, as it is difficult to tune the loss accordingly.

Also, (Q4) is thus answered affirmatively by this experiment as it overall shows that XIL can ``cure'' already confounded models, which is an important property for real-world applications.

\section*{Discussion}
The previous sections demonstrated that modifying XIL modules is no free lunch in the sense that modifying one module does not guarantee improvements in all criteria. In the following, we wish to discuss some additional points.

\begin{figure}[t]
    \centering
    \includegraphics[width=0.5\textwidth]{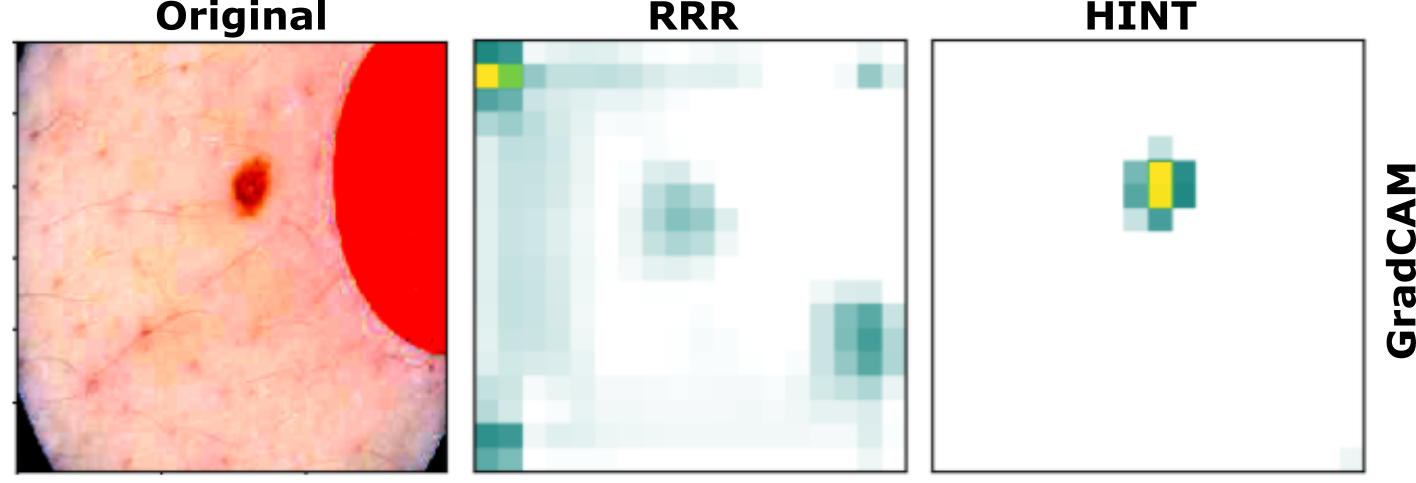}
    \caption{(left) An ISIC19 image with confounder (red patch). (middle) an RRR-revised model and (right) a HINT-revised model generate explanations for the image. The explanations are visualized with GradCAM. RRR helps discover yet unknown confounders (dark corners), and HINT reveals the potential of the reward strategy.}
    \label{fig:isic_showcase}
\end{figure}

As pointed out initially, it is often easier to state a wrong reason than a right reason \cite{Schramowski2020_Plantphenotyping}. However, penalizing wrong reasons may not be enough to revise a model to be right for \textit{all} the right reasons. Avoiding one wrong reason, but using another wrong one instead is not desirable.
The provided attribution maps for ISIC19 (cf. Fig.~\ref{fig:isic_showcase}) illustrate this trade-off. As we can observe from the attribution maps, the reward strategy (HINT) visually outperforms the penalty strategy. The penalty method, exemplified here via RRR, does to a certain degree point towards the right reason, but not as reliably as rewarding via HINT.
In general, however, a reward strategy cannot guarantee that confounders are avoided.

Throughout our work, we encountered ambiguities between different explainer methods. When a XIL method is applied and a sample is visualized with a different explainer method, than was optimized with, we find contradicting attribution maps (cf. RRR columns in Fig.~\ref{fig:MNISTheatmap}). In fact, the analysis of attribution maps shows remarkable differences between IG, GradCAM, and LIME. In some cases, we can even observe opposing explanations. Moreover, the GradCAM explanation for the Vanilla model in Fig.~\ref{fig:MNISTheatmap}a does not show a confounder activation although the scores in Tab.~\ref{tab:exp_performance_overall} clearly pinpoint a shortcut behavior of the model.
This consequently raises concerns about how reliable and faithful the explainer methods are. At this point, we note that for investigating XIL, we make the general assumption about the correctness of the explanation methods, which is a yet open topic in the field \cite{adebayo2018sanity, krishna2022disagreement}. Although it is the explicit goal of XIL to improve explanations, this can only work if an explainer method does not inherently fail at producing meaningful and coherent explanations. In that case, the overall objective of increasing user trust is already undermined before XIL enters the game. 
One of the main challenges of XIL is the real-world application. Revising a model must be easy for an average ML practitioner or any human user. If the resource demand is too high, the methods are difficult to use. This is specifically a problem for state-of-the-art, large-scale pre-trained models. One example is RBR which uses IFs, i.e. second-order derivatives, to correct Clever-Hans behavior. In our evaluations, we found that IFs induce a huge resource demand, making XIL slower and more challenging to optimize --loss clipping was necessary to avoid exploding gradients.

In terms of architecture choice and design, we also encountered several obstacles. Our typology description already pointed out that not every XIL method is applicable to every model or explainer method, e.g. GradCAM-based XIL methods can only be applied to CNNs. We argue a flexible XIL method is preferable such that various models and explainer methods can be applied.

From our experimental evaluations considering the number of required interactions, we observed that CE, with the \textit{dataset} augmentation strategy, requires the largest amount of user feedback. Especially for large-scale models, the number of interactions required can be a limiting factor. In practical use cases, often only a limiting number of explanatory feedback is available. Another aspect here is the aspect of trustworthiness in that a user might not trust a model as much if the feedback they have provided is not directly incorporated by the model and should suffice to revise a model. 
Furthermore, we noticed that CE is less robust to incomplete feedback, possibly compromising this approach alone for real-world application. Hence, combining CE with a loss-based XIL approach could be advantageous.

Lastly, we wish to note that a very noteworthy potential of XIL could be observed in the qualitative evaluations of ISIC19 attribution maps. In fact, by applying XIL on one confounder, we could identify further yet unknown confounders (shortcuts) to the user, in this case, the dark corners found in the images (cf. Fig.~\ref{fig:isic_showcase} (middle)). These findings further demonstrate the importance of a strong level of human-machine interactions via explanations. Particularly in such a setting, each can learn from the other in a stronger bidirectional, discourse-like manner and more than just the unidirectional way of communication provided by XAI alone.
To this end, we refer to the theory of embodied intelligence, in which interaction and manipulation of the environment allow for information enrichment to obtain intelligent systems \cite{Tan2007IntelligenceTI}.

\section*{Conclusions}
In summary, this work presents a comprehensive study in the rising field of XIL. Specifically, we have proposed the first XIL typology to unify terminology and categorize XIL methods concisely. Based on this typology we have introduced novel benchmarking tasks, each targeting specific aspects of the typology, for properly evaluating XIL methods beyond common accuracy measures.  
These cover the performance in model revision, robustness under changing feedback quality, interaction efficiency, and real-world applicability. In addition, we introduced a novel \wrmeasure measure for our experiments to quantify an average confounder activation in a model's explanations. Lastly, we showcased the typology and novel benchmark criteria in an empirical comparison of six currently proposed XIL methods.

Our typology and evaluations showed that XIL methods allow one to revise a model not only in terms of accuracy but also explanations.
However, we also observed overfitting to the individual explainer method being used. Avenues for future research are a mixture of explainers that may account for uncertainty on the right reasons.
Moreover, one should combine feedback on what the right explanation is with feedback on what it is not, rather than focusing on only one of these two feedback semantics. The weighting of each feedback based on the knowledge and feedback certainty of the user goes one step further.
And, of course, application to large-scale pre-trained models is an exciting avenue for future work.
Most importantly, existing XIL approaches just follow a linear \Select, \Explain, \Obtain, and \Revise. As in daily human-to-human communication, machines should also follow more flexible policies such as an \Explain\xspace \& \Obtain \xspace sub-loop, pushing for what might be called explanatory cooperative AI~\cite{Dafoe2021nature}. 

Overall, in this work, apart from a typology we have also introduced a set of novel measures and benchmarking tasks for differentiating and evaluating current and future XIL methods. This toolbox is by no means conclusive and should act as a starting point for a more standardized means of evaluation. Additional or improved measures and tasks should be investigated in further research, e.g. the benefits of mutual information.

\section*{Methods}
In this section, we present existing XIL methods and the measures and benchmarks we use and at the same time propose to evaluate.
\subsection*{XIL Methods}\label{xil_methods}
The fundamental task of XIL is to integrate the user's feedback on the model's explanations to revise its learning process. To tackle this core task, recently, several XIL methods have been proposed. 
Below we describe these methods in detail, dividing them based on two revision strategies: revising via (1) a loss term or (2) dataset augmentation. Both strategies rely on local explanations. 

\subsubsection*{Loss Augmentation}\label{loss_augmentation}
Strategy (1) can be summarized as optimizing Eqn.~\ref{eq:loss_aug} \cite{Rieger2019_CDEP}, where $X$ denotes the input, $y$ ground truth labels and $f$ a model parameterized by $\theta$. We optimize 
\begin{equation}\label{eq:loss_aug}
\min_{\theta}
\underbrace{ L_{\text{pred}}(f_{\theta}(X), y)}_\text{Prediction error} + \underbrace{\lambda\,  L_{\text{exp}}(\text{expl}_{\theta}(X), \text{expl}_{X})}_{\text{Explanation error}} \quad ,
\end{equation}
where $L_{\text{pred}}$ is a standard prediction loss, e.g. cross-entropy, guiding the model to predict the right answers, whereas $L_\text{exp}$ ensures the right reasons, i.e. right explanations, scaled by the regularization rate $\lambda$.

\paragraph{Right for the Right Reasons (RRR)}
In the work of Ross \etal~\cite{Ross2017_RRR}, the objective is to train a differentiable model to be right for the right reason by explicitly penalizing wrong reasons, i.e. irrelevant components in the explanation. That means \Revise \xspace enforces a penalty strategy. To this end, this approach generates gradient-based explanations $\text{expl}_{\theta}(X)$ and restricts them by constraining gradients of irrelevant parts of the input. For a model \mbox{$f(X|\theta) = \hat{y} \!\in\! \mathbb{R}^{N \times K}$} and inputs \mbox{$X\!\in\!\mathbb{R}^{N \times D}$} we get
\begin{equation}
L_{\text{exp}} = \sum_{n=1}^N \left(M_n \, \text{expl}_{\theta}(X_{n})\right)^2,
\end{equation}
where N is the number of observations, K is the number of classes, and D is the dimension of the input.
With this loss term, the user's explanation feedback \mbox{$M_n \!=\! \text{expl}_X$}, indicating which input regions are irrelevant, is propagated back to the model in the optimization phase. The loss prevents the model from focusing on the masked region by penalizing large values in this region. According to the authors, \mbox{$L_{\text{pred}}$} and \mbox{$L_{\text{exp}}$} should have the same order of magnitude by setting a suitable regularization rate $\lambda$ in Eqn.~\ref{eq:loss_aug}. 

Ross \etal~\cite{Ross2017_RRR} implement \Explain \xspace with IG by generating explanations based on first-order derivatives, i.e. \mbox{$\text{expl}_{\theta}(X)\!=\!I\!G(X)$}.
However, RRR's \Explain \xspace is not limited to this explainer. Schramowski \etal~\cite{Schramowski2020_Plantphenotyping} propose Right for the Right Reason GradCAM (\textbf{RRR-G}) generating explanations via \mbox{$\text{expl}_{\theta}(X)\!=\!GradC\!AM(X)$} and Shao \etal~\cite{Shao2021_RBR} propose Right for the Better Reasons (\textbf{RBR}) with second-order derivatives, i.e. \mbox{$\text{expl}_{\theta}(X)\!=\!I\!F(X)$}. We describe further mathematical details in Supplement~D.
In order to penalize wrong reasons, \Obtain \xspace in this case expects feedback in the following form. A user annotation mask is given as \mbox{$\text{expl}_X = M \!\in\! \{0, 1\}^{N \times D}$} with $1$s indicating wrong reasons. E.g. in this work, we consider confounding pixels as wrong reasons.

\paragraph{Contextual Decomposition Explanation Penalization (CDEP)}
Compared to the others, CDEP \cite{Rieger2019_CDEP} uses a different explainer method, CD, i.e. its \Explain \xspace module is restricted to this explainer method, \mbox{$\text{expl}_{\theta}(X)\!=\!C\!D(X)$}. The CD algorithm measures the layer-wise attribution of a marked feature, here image region, to the output. It decomposes the influence on the prediction between the marked image region to the remaining image. This enables only focusing on the influence of the marked image region and, in this case, penalizing it. Hence, \Revise \xspace is implemented again with the penalty strategy. The user mask $M$ penalizes the model explanation via
\begin{equation}
L_{\text{exp}} = \sum_{n=1}^N \left\|\text{expl}_{\theta}(X_{n}) - M_n\right\|_1.
\end{equation}

\paragraph{Human Importance-aware Network Tuning (HINT)}
% \textbf{Human Importance-aware Network Tuning (HINT).}
In contrast to previous methods, HINT \cite{Selvaraju2019_HINT} explicitly teaches a model to focus on \textit{right reasons} instead of \textit{not} focusing on \textit{wrong reasons}. In other words, HINT rewards activation in regions on which to base the prediction, whereas the previous methods penalize activation in regions on which \textit{not} to base the prediction. Thus, \Revise \xspace is carried out with the reward strategy.
\Explain \xspace can take any gradient-based explainer, whereas the authors implemented it with GradCAM, i.e. $\text{expl}_{\theta}(X)\!=\!GradC\!AM(X)$. Finally, a distance, e.g. via mean squared error, is computed between the network importance score, i.e. generated explanation, and the user annotation mask, resulting in:
\begin{equation}
L_{\text{exp}} = \frac{1}{N} \sum_{n=1}^N \left( \text{expl}_{\theta}(X_{n}) - M_{n} \right)^2.
\end{equation}
Importantly, \Obtain \xspace differs from previous methods in that $1$s in the user annotation mask $M$ mark right reasons, not wrong reasons. We define relevant pixels (components) as the right reasons for our evaluations.

\subsubsection*{Dataset Augmentation}
In contrast to the XIL methods, which add a loss term to revise the model, i.e. to implement \Revise, further XIL methods exist which augment the training dataset by adding new (counter)examples to the training data \cite{Teso2019_CAIPI}. Where the previous approaches directly influence the model's internal representations, this approach indirectly revises a model by forcing it to generalize to additional training examples, specifically tailored to remove wrong features of the input space. This augmentation can, e.g., help remove a model from focusing on confounding shortcuts.

\paragraph{CounterExamples (CE)}
Teso and Kersting \cite{Teso2019_CAIPI} introduce CE, a method where users can mark the confounder, i.e. wrong reason, region in an image from the training data and add a corrected image, i.e. in which an identified confounder is removed, to the training data.

In comparison to strategy (1), this strategy is model- and explainer-agnostic, i.e. \Explain \xspace can be implemented with any explainer method as user feedback is not processed directly via the model's explanations. Specifically, \Obtain \xspace takes user annotation masks that mark the components in the explanation that are incorrectly considered relevant. In this case, the explanation corrections are defined by $\mathcal{C}=\{j:|w_j|> 0  \wedge j$th component marked by user as irrelevant\}, where $w_j$ denotes the $j$th weight component in the attribution map. These explanation corrections are transformed into counterexamples in order to make the feedback applicable to the model. A counterexample is defined as \mbox{$j\in\mathcal{C}: \{(\overline{X}, \overline{y})\}$}, where $\overline{y}_i$ is the, if needed, corrected label and $\overline{X}_i$ is the identical input, except the previously marked component. This component is either (1) randomized, (2) changed to an alternative value, or (3) substituted with the value of the $j$th component appearing in other training examples of the same class. The counterexamples are added to the training dataset. Moreover, it is also possible to provide multiple counterexamples per correction, e.g., different strategies. In our case, where the input is an image, the user's explanation correction is a binary mask, and a counterexample is an original image with the marked pixels being corrected. 

Instead of using noise to augment an example, Lang \etal~\cite{Lang2021ExplainingIS} present an attractive alternative that generates new realistic examples from a style space learned with a GAN-based approach.

\subsection*{Evaluating XIL is More Than Just Accuracy}
Although a variety of works on XIL exist, there remains a research gap due to the lack of an in-depth comparison between these. Moreover, XIL methods are, if at all, usually only compared %to RRR 
in terms of accuracy on benchmark confounded datasets.
This essentially only measures if a XIL method successfully %succeeds in revising the model, e.g. 
helps overcome the confounder in terms of predictive power. However, the goal of XIL goes beyond overcoming confounders and also includes improving explanations overall, e.g. outside of a confounding setting. Hence, a profound examination that focuses on different aspects of the typology is crucial for a sound analysis of current and future research and for filling this research gap. We, therefore, extend our typology by proposing additional measures and benchmarking tasks for a thorough evaluation of a XIL method and clarify these in the following sections. 

\subsubsection*{Measures for Benchmarking}
In the following, we present existing and introduce novel quantitative and qualitative approaches to evaluate XIL methods.

\paragraph{Accuracy}
Many previous works on XIL revert to measuring prediction accuracy as a standard measure to evaluate a method's performance. This mainly works by using a confounded dataset in which the predictive accuracy on a non-confounded test set serves as a proxy for ``right reasons''. However, this measure can only be used to evaluate XIL on datasets with known confounders and test sets that do not contain the confounding factor. Otherwise, yet unknown confounders may still fool the model and prevent an accurate evaluation of a XIL method. This is particularly important as XIL does not only aim to improve the predictive power, but also to improve the quality of the model's explanations regarding the preferences and knowledge of the human user. We note that in all further mentions of ``test accuracy'' we are considering a model's performance based on the data set classification rate on an unseen test set.

\paragraph{Qualitative Explanation Analysis}
Another approach to evaluating the effectiveness of XIL methods is to qualitatively inspect a model's explanations, e.g. attribution maps, prior to and after revisions. Next to the previously mentioned test accuracy, this approach to quality assessment is another popular measure many previous works focus their evaluations on. Some recent techniques for (semi-)automatic explanation analysis exist, e.g. for detecting Clever-Hans strategies. For example, Spectral Relevance Analysis (SpRAy) inspects and clusters similar explanations \cite{Lapuschkin2019CleverHans,Anders2019}. 

\paragraph{Wrong Reason Measure}
Besides standard measures like accuracy, we, therefore, propose a novel, yet intuitive measure, termed \textit{wrong reason} measure (\wrnospace), to measure how wrong a model's explanation for a specific prediction is, given ground-truth (user-defined) wrong reasons. In contrast to the qualitative evaluation (manual, visual inspection) of attribution maps, our \wrmeasure measure provides a quantitative complement.

In detail, given an input sample $X$, e.g. an image, a model $f$ with parameters $\theta$, an explainer $expl$ and ground-truth annotation mask $M$, we quantify \wrmeasure as 
\begin{equation}
    \wrnospace(X, M) =  \frac{sum(b_{\alpha}(norm^+(expl_{\theta}(X))) \circ M)}{sum(M)} ,
\end{equation}
where $\circ$ is the Hadamard product, and $norm^+$ normalizes the attribution values of the explanation to $[0,1]$, while only taking positive values into account by setting all negative values to zero.
$b_{\alpha}$ binarizes the explanation (\mbox{$expl_{ij} > \alpha \Rightarrow 1$ else $0$}) and the threshold $\alpha$ can be determined by the pixel-level mean of all explanation attribution maps in the test set beforehand.

Depending on the explainer $expl$, it might be necessary to scale (down/up) the dimensions of the explanation to match dimension $d$ of the annotation mask $M$. 
In short, the measure calculates to what extent the \textit{wrong reason} area is activated.
$\wrnospace\!=\!1$ translates to $100\%$ activation of the \textit{wrong reason} region, indicating that the model gets fooled by the \textit{wrong reason} and spuriously uses it as an informative feature. If this behavior co-occurs with high predictive performance this will imply Clever-Hans behavior and reasoning based on a \textit{wrong} reason. Contrarily, $\wrnospace\!=\!0$ signals that $0\%$ of the \textit{wrong reason} area is activated. However, it is worth noting that one cannot, in principle, claim that the model's reasoning is based on the \textit{right} reason from being \textit{not wrong}.

Comparing the \wrmeasure scores of a Vanilla model with a XIL-extended model allows us to estimate the effectiveness of a specific XIL method. As one objective of XIL is to overcome the influence of the \textit{wrong reason} area, the \wrmeasure score should be at least smaller than the score for the Vanilla model. 

\subsubsection*{Novel Benchmarking Tasks}
In the following, we introduce further relevant benchmarking tasks for evaluating XIL methods.

\paragraph{Feedback Robustness}
An important aspect of the usability of a XIL approach is its robustness to the completeness of and quality variations within the user feedback. This task is vital, as user feedback in the real world is error-prone.
In order to provide a benchmark that is comparable between different datasets and can be efficiently evaluated, we propose to simulate and model robustness via a proxy task for all dataset-model combinations. In the spirit of Doshi-Velez and Kim \cite{Doshi-Velez2017_eval}, this task is, therefore, a functionally-grounded evaluation, with no human in the loop. 

Two compelling cases to examine are cases of arbitrary and incomplete feedback. Arbitrary feedback can also be viewed as a sanity check of a XIL method since it should not change the performance in any direction. In other words, a model should not produce worse or better predictive performance, as the feedback is neither useful nor detrimental. On the other hand, incomplete feedback imitates real-world feedback by providing only partially valuable feedback.
For instance, in the case of the DecoyMNIST (for details see \textit{Experimental Setup}), two scenarios can be modeled as follows:
\begin{enumerate}
    \item \textit{Arbitrary} feedback: $5\!\times\!3$ rectangle pixel region in the middle sections of the top or bottom rows of an image, thus neither on relevant digit feature regions nor on any parts of confounder squares, i.e. $M\!\neq\!C$
    \item \textit{Incomplete} feedback: with subregion $S$ (here top half) of relevant components $C$. Thus, $ M = \mathbf{1}_S\!:\!C$
\end{enumerate}
A feedback mask is again denoted by $M$ and the set of (ir)relevant components by $C$. In the case of correct user feedback $M\!=\!C$.

CE uses manipulated copies of the original images instead of binary masks. There are different CE strategies to manipulate (we chose CE-strategy \textit{randomize}). The manipulated images are added to the training set. Exemplary feedback masks for this experiment are illustrated in Supplement Fig.~1, visualizing the feedback types and further details and examples can be found in Supplement~B.1.

\paragraph{Interaction Efficiency}
In many previous applications and evaluations of XIL methods, every training sample was accompanied by corresponding explanatory feedback. Unfortunately, feedback, e.g. in the form of input masks, can be costly to obtain and potentially only available to a limited extent. 
A very relevant evaluation, particularly for a method's practical usability, is how many feedback interactions are required for a human user to revise a model via a specific XIL method. In other words, we propose to investigate the \textit{interaction efficiency} of a method as an additional benchmark task.

To simulate a reduced feedback size, we propose to randomly sample a subset of $k$ annotations before training and evaluate each model with the different-sized feedback sets. Different values for $k$, i.e. number of explanatory interactions, enable a broad insight into the capability of a XIL method to revise a model efficiently. The effect of the reduced set size is measured with accuracy. Thus, this evaluation task reduces the feedback set size and investigates its impact on the overall effectiveness of a XIL method. 

\paragraph{Switch XIL on}
A further benchmark task that we propose is called \textit{switch XIL on} and is motivated in two ways. First, it complements previous works that often only simulated interaction with a model from scratch but not from a strongly confounded model, which would grant more insight into the effectiveness and function of XIL. Second, Algorithm~\ref{alg:typology} shows that a model is usually fitted to the given data beforehand, and XIL is applied to the confounded model after e.g. Clever-Hans behavior is detected. This contrasts with the other evaluation tasks, where often a model is optimized via a XIL method from scratch. In addition to related work investigating the real-world applicability of XIL with a more real-world dataset \cite{Slany2022CAIPIIP}, we want to propose methods for the same purpose instead. This property is essential, as completely retraining a model can be very costly or even infeasible, e.g. for large-scale pre-trained models. Hence, it would be very valuable for a XIL method if it can successfully be applied in revising an already corrupted model.

This evaluation task targets correcting a pre-trained, strongly corrupted model, i.e. a model already strongly biased towards Clever-Hans behavior. 
To this end, a Vanilla model is trained on the confounded train set for several epochs. Subsequently, the XIL loss is switched on (for CE, the train set is augmented).

\subsection*{Experimental Protocol}
For our experiments, we use two different models: a simple CNN for the benchmark and a VGG16 for the scientific dataset. We use RRR with different explainer methods (IG (RRR), GradCAM (RRR-G), and IF (RBR)) to not only compare different XIL methods but also to investigate the impact of different explainer methods on the same XIL method. For simplicity, we only investigate one XIL method with different explainers (Tab.~\ref{tab:xil_summary}).
In this work, we optimize our models with Adam \cite{kingma2014_adam} and a learning rate of $0.001$ for $50$ epochs. For the standard experiments, the right reason loss is applied from the beginning. For Decoy(F)MNIST, we use the standard train-test split ($60000$ to $10000$) and for the ISIC19 dataset, we use a $80-20$ split. We set the batch size to $256$ for the benchmark datasets and to $16$ for ISIC19.
Further experimental details can be found in Supplement~A and B.

The \textbf{DecoyMNIST} dataset \cite{Ross2017_RRR} is a modified version of the MNIST dataset, where the train set introduces decoy squares. 
Specifically, train images contain $4\!\times\!4$ gray squares in randomly chosen corners, whose shades are functions of their digits. These gray-scale colors are randomized in the test set. The binary feedback masks $M$ mark confounders for the penalty strategy, while the masks mark the digits for the reward strategy.

FashionMNIST (FMNIST) is an emendation of MNIST, as it is overused in research and limited in complexity. FMNIST consists of images from ten fashion article classes. The \textbf{DecoyFMNIST} dataset introduces the same confounding squares as DecoyMNIST.

The \textbf{ISIC (International Skin Imaging Collaboration) Skin Cancer 2019} dataset \cite{Codella2017, Combalia2019, Tschandl2018} consists of high-resolution dermoscopic images of skin lesions, having either a benign or malignant cancer diagnosis. In contrast to the benchmark datasets and in addition to related work on a medical toy dataset \cite{Slany2022CAIPIIP}, this dataset is significantly more complex and covers a real-world high-stakes scenario. The main difference is that the confounders are not added artificially, and we only know of one confounder, while there can still exist unknown confounders. The known confounders are colored patches next to a skin lesion.
We adjust the original test set as it contains images with both known and unknown confounders. We exclude the images with the known confounder (patches) to ensure a more non-confounded test set, which is essential to measure the confounder influence. Note, the dataset only contains images of Europeans with lighter skin tones, representing the well-known skin color problem, and therefore results cannot be generalized to other skin tones.

\section*{Data availability}
All datasets are publicly available. We have two benchmark datasets, in which we add a confounder/ shortcut, i.e. we adjust the original dataset, and a scientific dataset where the confounder is not artificially added but inherently present in the images. The MNIST dataset is available at \url{http://yann.lecun.com/exdb/mnist/} and the code to generate its decoy version at \url{https://github.com/dtak/rrr/blob/master/experiments/Decoy\%20MNIST.ipynb}. The FashionMNIST dataset is available at \url{https://github.com/zalandoresearch/fashion-mnist} and the code to generate its decoy version at \url{https://github.com/ml-research/A-Typology-to-Explore-the-Mitigation-of-Shortcut-Behavior/blob/main/data_store/rawdata/load_decoy_mnist.py}. The scientific ISIC dataset with its segmentation masks to highlight the confounders are both available at \url{https://isic-archive.com/api/v1/}. 

\section*{Code availability}
All the code \cite{github-link} to reproduce the figures and results of this article can be found at \url{https://github.com/ml-research/A-Typology-to-Explore-the-Mitigation-of-Shortcut-Behavior} (archived at \url{https://doi.org/10.5281/zenodo.6781501}). The CD Algorithm is implemented at \url{https://github.com/csinva/hierarchical-dnn-interpretations}. Furthermore, other implementations of the evaluated XIL algorithms can be found in the following repositories: RRR at \url{https://github.com/dtak/rrr}, CDEP at \url{https://github.com/laura-rieger/deep-explanation-penalization}, and CE at \url{https://github.com/stefanoteso/calimocho}.

\section*{Acknowledgments}
The authors thank the anonymous reviewers for their valuable feedback. Furthermore, the authors thank Lars Meister for the preliminary results and insights on this research. This work benefited from the Hessian Ministry of Science and the Arts (HMWK) projects ‘The Third Wave of Artificial Intelligence—3AI’, hessian.AI (F.F., W.S., P.S., K.K.) and ‘The Adaptive Mind’ (K.K.), the ICT-48 Network of AI Research Excellence Centre ‘TAILOR’ (EU Horizon 2020, GA No 952215) (K.K.), the Hessian research priority program LOEWE within the project WhiteBox (K.K.), and from the German Center for Artificial Intelligence (DFKI) project ‘SAINT’ (P.S., K.K.).

\section*{Author Contributions Statement}
FF, WS, and PS designed the experiments. FF conducted the experiments.
FF, WS, PS, and KK interpreted the data and drafted the manuscript.
KK directed the research and gave initial input. 
All authors read and approved the final manuscript.

\section*{Competing Interests Statement}
The authors declare no competing interests.

\appendix
\section*{\centering Supplementary Information for \\ A Typology to Explore the Mitigation of Shortcut Behavior}

\begin{figure}[b]
  \centering
    \begin{subfigure}{0.48\textwidth}
  \centering
        \includegraphics[width=0.6\linewidth]{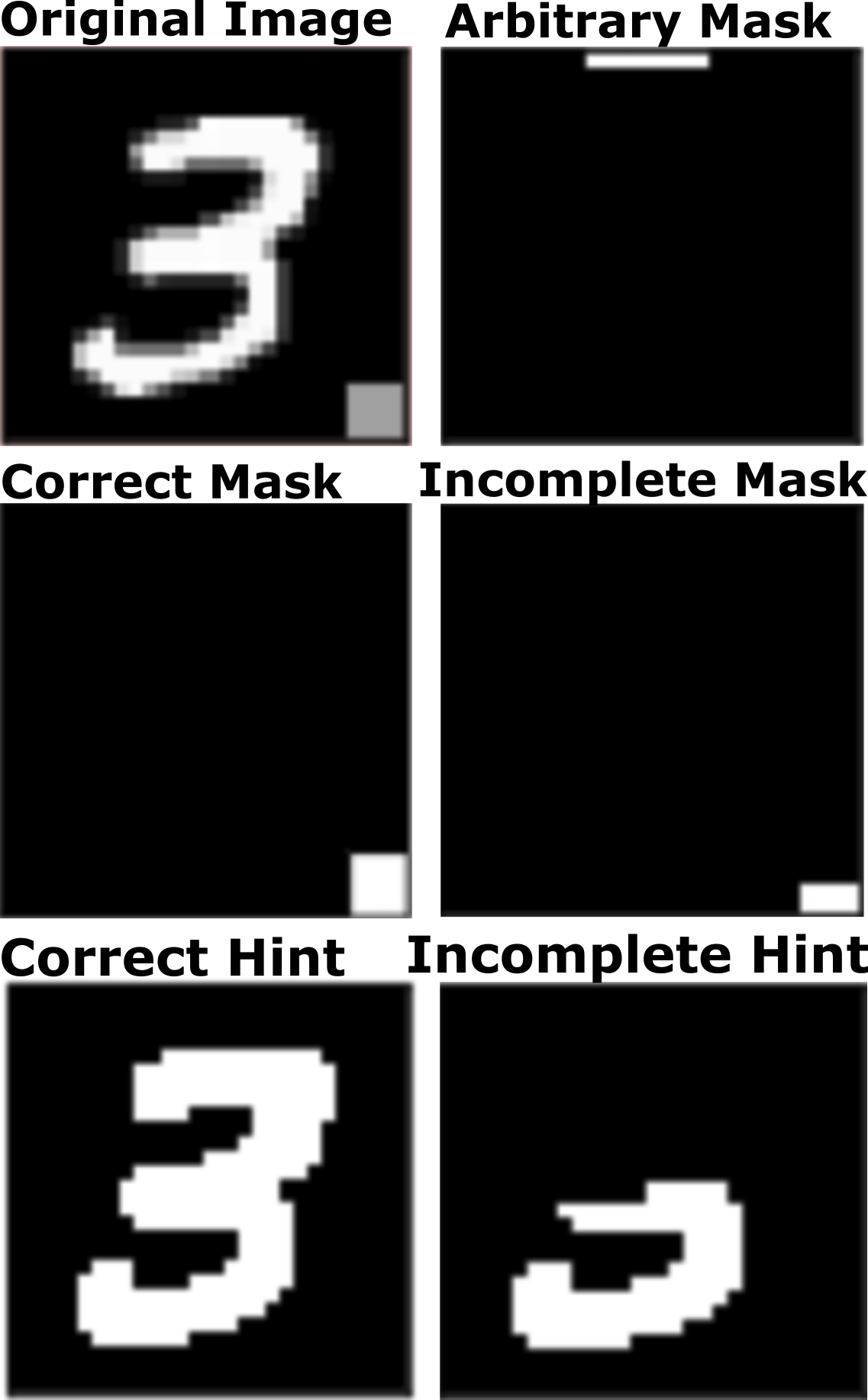}
        \caption{Counter examples}
      \label{fig:int_eff_examples_mask}
    \end{subfigure}
    \begin{subfigure}{0.48\textwidth}
      \centering
        \includegraphics[width=0.8\linewidth]{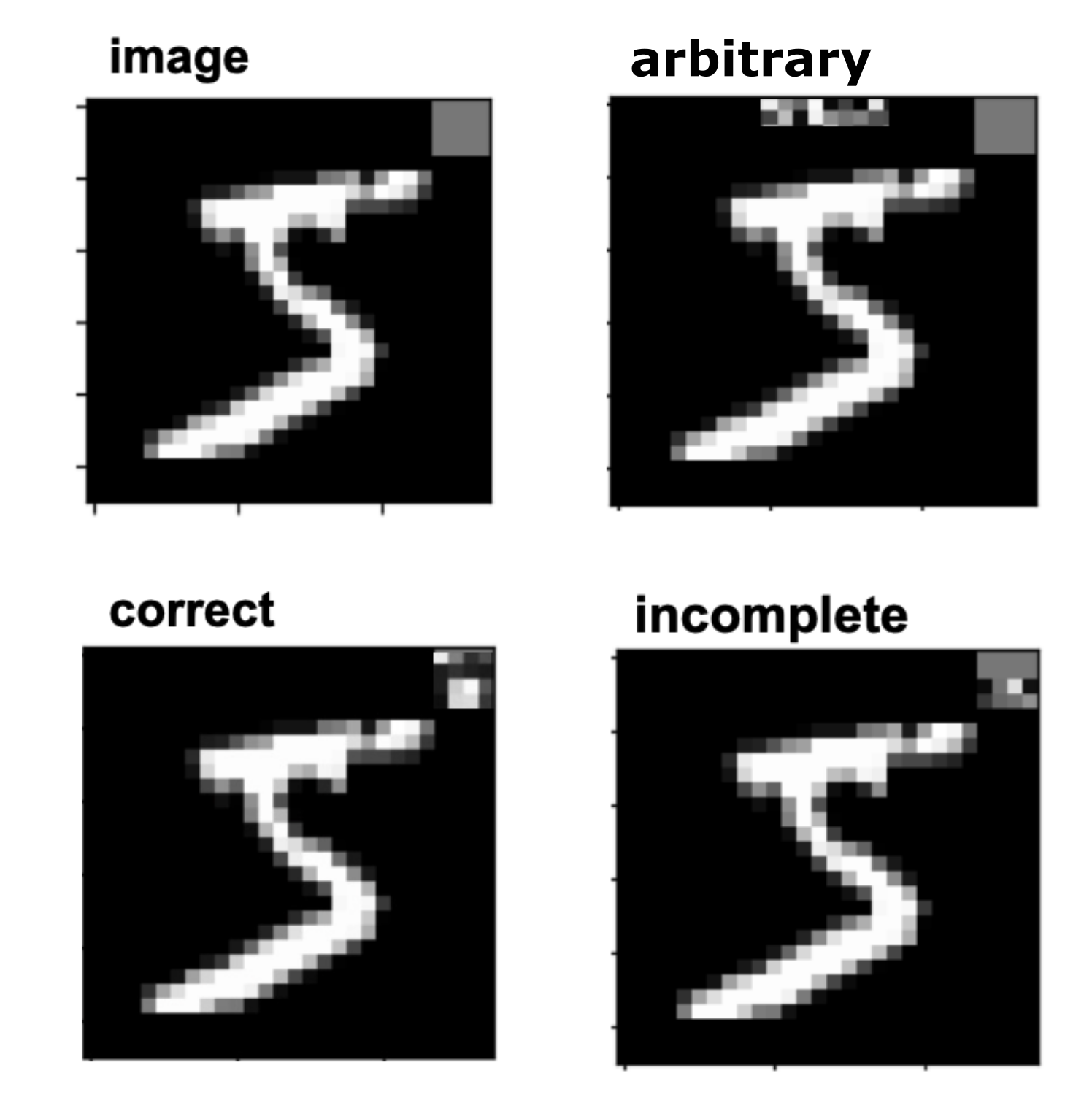}
        \caption{Attribution masks}
    \label{fig:int_eff_examples_ce}
    \end{subfigure}
    \caption{(a) attribution masks and (b) counter examples (with strategy \textit{randomize}) for an DecoyMNIST example. (a) The original images, here the digits ``3'' and ``5'', have a confounder, here a gray decoy square in the corner. Below the original images, there are a correct feedback masks (counter examples) for penalizing the wrong reason and a correct feedback mask/hint for rewarding the right reason. The right side shows arbitrary and incomplete feedback masks (counter examples) for the feedback robustness experiment.}
  \label{fig:int_eff_examples}
\end{figure}

\subsection*{A \; Used Models and Datasets}\label{used_models} 

For the benchmark datasets, we ran the experiments on a CNN consisting of two convolution layers (channels=[20,50], kernel size=5, stride=1, pad=0), each followed by a ReLU activation and max-pooling layer. The last two layers are fully-connected.    

Due to the high complexity of the ISIC19 dataset, we decided to use a more sophisticated architecture; the popular VGG16 model \cite{Simonyan2014_VGG}. It is commonly used across the CV community and established to evaluate CV tasks. More precisely, we used a VGG16 model, which was pre-trained on ImageNet \cite{deng2009imagenet}. We used the available VGG16 of the PyTorch library \cite{pytorch}; see docs at \url{https://pytorch.org/vision/stable/models.html}. Hence, we normalize the RGB channels of the images to the expected range of $mean = [0.485, 0.456, 0.406]$ with $std = [0.229, 0.224, 0.225]$ (i.e. based on ImageNet training set).
We provide an overview of the datasets used and a mapping of the dataset and model in Tab.~\ref{tab:used_datasets}.

\begin{table*}[h]
    \centering
    \begin{tabular}{c|lccccccc}
    Dataset & Size & split & Classes & Input-dim & Confounder & Type & Source & Model\\ \hline
    DecoyMNIST & 70k & 6:1 & 10 & 28×28×1 & gray squares & benchmark & \cite{Ross2017_RRR} & S-CNN\\ 
    DecoyFMNIST & 70k & 6:1 & 10 & 28×28×1 & gray squares & benchmark & \cite{xiao2017_FashionMnist}  & S-CNN\\ 
    ISIC Skin Cancer 2019 & 21k & 8:2 & 2 & 650×450×3 & colored patches & scientific & \cite{Codella2017, Combalia2019, Tschandl2018}  & VGG16\\
    \end{tabular}
    \caption{Overview of used datasets.}
    \label{tab:used_datasets}
\end{table*}
\begin{table}[t]
    \centering
    \begin{tabular}{c|cccccc}
         & RRR & RRR-G & RBR & CDEP & HINT & CE \\
         \hline
         DecoyMNIST & 10 & 1 & 100000 & 1000000 & 1000 & --  \\
         DecoyFMNIST & 100 & 1 & 100000 & 1000000 & 0.1 & --
    \end{tabular}
    \caption{Regularization values $\lambda$ used for all experiments. Values derived from the hyperparameter grid search in Fig.~\ref{fig:reg-rates} for the Decoy(F)MNIST dataset. CE does not use the loss augmentation strategy and hence has no regularization value.}
    \label{tab:lambda_vals}
\end{table}

\subsection*{B \; Details on Experimental Setup}
We provide further details on the experimental setup to make our results reproducible. 

\subsubsection*{B.1 \; Feedback Quality Examples}\label{sec:feedback_examples}
In Fig.~\ref{fig:int_eff_examples_mask}, we show exemplary feedback masks. The figure describes correct, arbitrary, and incomplete feedback for an example of the DecoyMNIST dataset. The image with digit ``3'' contains a decoy square in the corner as a confounder. Incomplete feedback uses the lower half of the correct annotation. We also illustrate the difference between reward (HINT) and penalty (others) feedback masks. Arbitrary feedback does not point to the confounder nor the right reason; it just targets non-relevant, i.e. arbitrary, regions. 

We further showcase the same for the CE methods in Fig.~\ref{fig:int_eff_examples_ce}. As CE uses the dataset augmentation strategy, the images shown are examples added to the datasets instead of masks incorporated in the loss. To create new (counter)examples, there are several strategies. We used and illustrate here the \textit{randomize} strategy. The figure also shows the different feedback qualities combined with the CE strategy.

\subsubsection*{B.2 \; Regularization rate stability.}
The regularization rate $\lambda$ is arguably the most crucial parameter of several XIL methods (loss-based methods), as it controls the influence of the right reason (RR) loss. As we already described in the main section, the RR loss guides the model to give the right explanations, i.e. base the predictions on the correct image regions, whereas the right answer (RA) loss directs the model to predict the right answers, i.e. predict the correct class. Note, CE does not augment the loss and therefore does not have a regularization rate.

The regularization rate needs to be set by the user/ ML practitioner for each method and each dataset individually. On real-world datasets, one does not necessarily have a particular non-confounded test set on which to tune hyperparameters. As we saw on the ISIC19 dataset (Tab.~\ref{tab:exp_performance_stds}), we were also faced with unclear test set performances. Moreover, in real-world scenarios, not necessarily all confounders are known beforehand. Thus, it is a desirable characteristic of a XIL method if the performance is relatively stable across a broader range of initialization values, as this robustness reduces the possibility of performance fluctuations.

To evaluate the stability of the XIL method wrt. the regularization rate $\lambda$, we trained the XIL-extended models on the Decoy(F)MNIST on a range of $\lambda$ values drawn from a logarithmic scale $[1e^{-2}, 1e^{6}]$. This also served as a grid search for the Decoy(F)MNIST experiment.

\begin{figure}[h]
  \centering
  \subfloat[DecoyMNIST]{\includegraphics[width=0.8\textwidth]{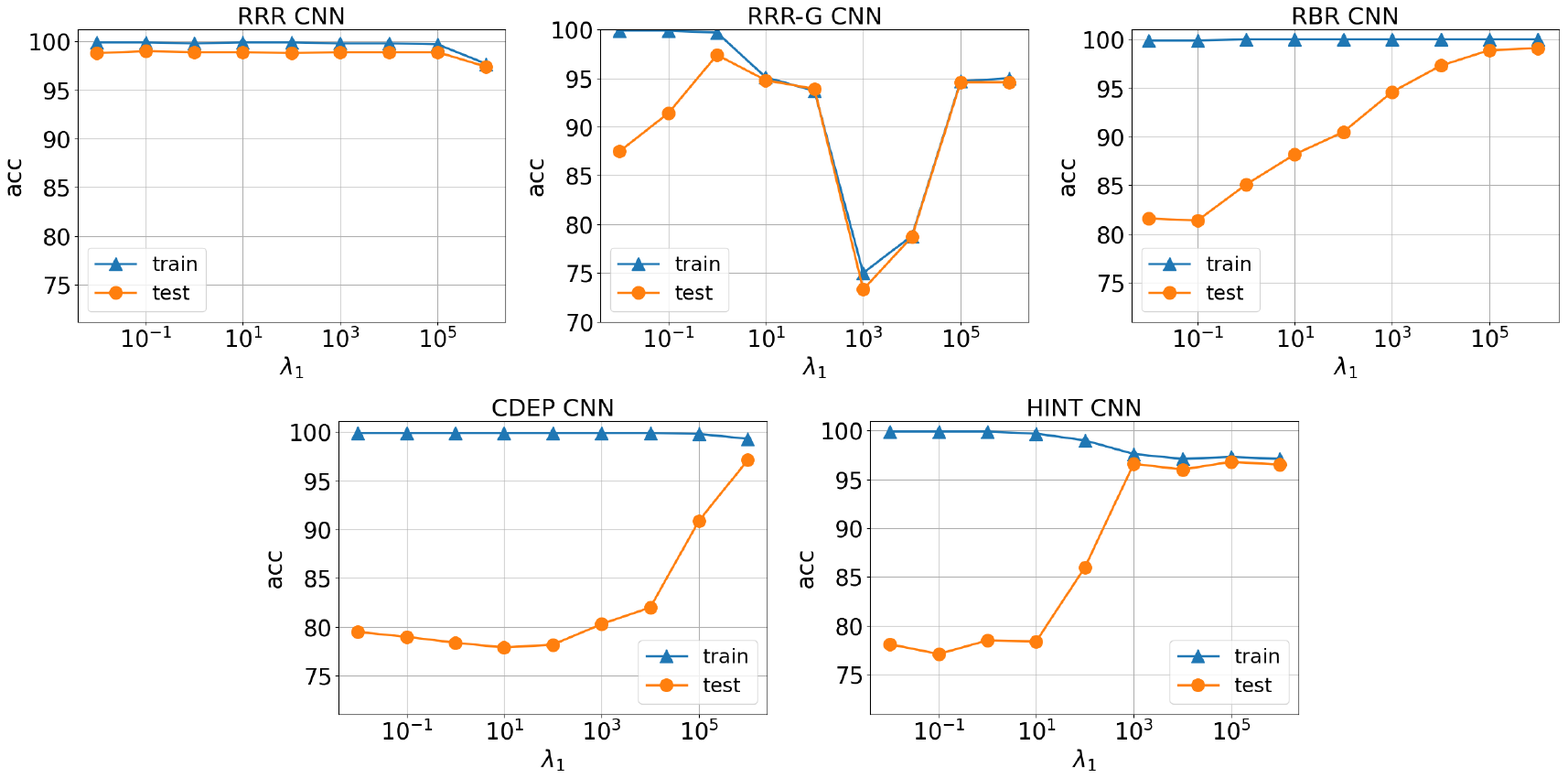}}
%   \caption[DecoyMNIST regularization rates stability]{DecoyMNIST regularization rates stability on S-CNN. Plots show the development of the test accuracy w.r.t. to different regularization rates $\lambda$ (logarithmic scale). We did not apply any technique that smoothes the training (RR-clipping); all runs were cross-validated on 5 different seeds [1, 10, 100, 1000, 10000]}
%   \label{fig:dm-reg-rates}
% \end{figure*}

% \begin{figure}[t]
%   \centering
  \subfloat[DecoyFMNIST]{\includegraphics[width=0.77\textwidth]{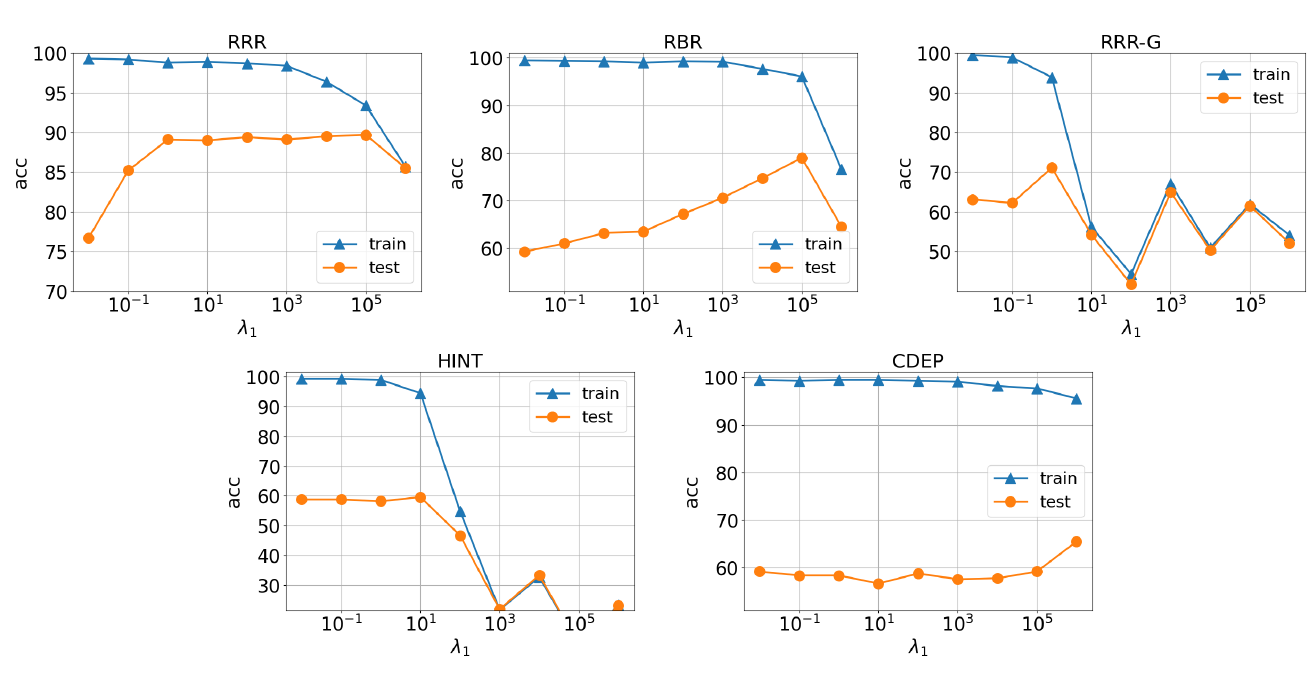}}
  \caption{Regularization rates stability experiment on (a) DecoyMNIST and (b) DecoyFMNIST with S-CNN. Plots show the development of the test accuracy w.r.t. to different regularization rates $\lambda$ (logarithmic scale); all models did not use any smoothing optimization techniques (RR-clipping); all runs were cross-validated on 5 different seeds [1, 10, 100, 1000, 10000]}
%   \caption[DecoyFMNIST regularization rates stability]{DecoyFMNIST regularization rates stability; All models did not use any smoothing optimization techniques; all runs are cross-validated on five different seeds [1, 10, 100, 1000, 10000]}
  \label{fig:reg-rates}
\end{figure}

As Fig.~\ref{fig:reg-rates} indicates, RRR was the most stable method across the evaluation interval. RBR and CDEP required relatively high rates, as their default ($\lambda\!=\!1$) RR loss was small. Setting $\lambda$ to a large value simultaneously increased the training instability, leading to more significant differences between individual loss updates per batch. If the RA and RR losses were far apart (e.g. Switch XIL ON experiments), we encountered training difficulties, hence we presume it to be helpful for them to be in the same order of magnitude. GradCAM-based XIL methods (RRR-G and HINT) required a smaller regularization rate, and bigger ones led to training instabilities. In general, we encountered occasional cases where the training would crash entirely, particularly RBR, HINT, and CDEP were susceptible. Training with those methods, we had to carefully calibrate the $\lambda$ value and applied an RR-loss clipping (see Appendix~B.3) to prevent the model from breaking down to random guessing. The final $\lambda$ values used for the experiments in this work are presented in Tab.~\ref{tab:lambda_vals}. Therefore, we picked the maximum value of the test performance from the grid search.

\begin{table}[t]
    \centering
    \begin{tabular}{cc|cccccc}
         & Vanilla & RRR & RRR-G & RBR & CDEP & HINT & CE \\
         \hline
         DecoyMNIST & 96 & 492 & 291 & 1009 & 451 & 286 & 178  \\
         DecoyFMNIST & 95 & 489 & 284 & 1205 & 445 & 341 & 179
    \end{tabular}
    \caption{Training time for each XIL method. The values are given in seconds and represent the average of 5 runs; lower is better.}
    \label{tab:resource_demand}
\end{table}

\subsubsection*{B.3 \; Resource Demand and Right Reason Loss Clipping}
Another interesting measure to compare different XIL methods for their computational efficiency. We provide the run times on both benchmark datasets in Tab.~\ref{tab:resource_demand} to get an insight into their relative performance. This experiment describes a standard learning procedure for $50$ epochs on an NVIDIA Tesla T4 GPU. As the table shows, the data augmentation technique (CE) simply doubles the run time as only the training data is doubled. The use of IFs (RBR) increases the run time compared to the Vanilla by nearly 13 times. GradCAM-based methods (RRR-G and HINT) are efficient loss-based XIL methods as GradCAM explanations are less resource-demanding. Depending on the task at hand, resource efficiency has to be considered a crucial factor for real-world application.

Throughout our evaluation, we encountered training instabilities mainly caused by the RA and RR loss imbalances, which most of the time led to exploding gradients. A straightforward strategy to ameliorate this problem is to clip the RR loss to a predefined maximum. The RR clipping enabled us to train XIL models with large regularization values $\lambda$. We used this mainly for RBR and RRR-G. Alternatively, one could also decrease the learning rate to enable smaller loss updates or use other techniques to better balance the RA and RR loss.

\subsection*{C \; Details on \wrmeasure Measure}
% \subsection*{C \; Details on Typology and \wrmeasure Measure}
% \subsubsection*{C.1 \; Classifying existing methods in.}

% \subsubsection*{C.1 \; Measuring \wrmeasure performance of XIL methods.}
For each trained model, we calculate the mean \wrmeasure score for all three explainer methods (IG, GradCAM, and LIME) over all test images containing a confounder. Images for which no explanation can be constructed are excluded. 
%For setting the threshold $\alpha$ of the binarization function $b_\alpha$, we determine the pixel-level mean over all explanation attribution maps in the test set beforehand. The binarization helps to deal with attribution maps containing large areas of zero activation or an imbalance in general.\ps{auch dieser paragraph hört sich eher nach experimenten an? Anders formulieren?}
% In contrast to the median - which was often zero or close to zero if more than half of the pixels of an explanation were not activated - using mean thresholds we were able to better incorporate intensity differences in the activation.
% \footnote{to illustrate: Imagine the case where more than half (51\%) the pixels are zero, 25\% region X is fully activated (1.0) and 24\% region C - which is the confounder - is marginal activated (0.1). We - as humans - would clearly label region X as most important and region C as unimportant for the prediction. Calculating median would yield a 0.0 threshold, which after applying the binarization would activate region C fully (all 1.0 now) and thus label it as highly important. On the contrast, calculating the mean threshold yields 0.274. But this time region C is not activated after the binarization and thus labeled as unimportant, which correlates with the human assessment.}.

Comparing the \wrmeasure scores of the Vanilla model with a XIL-extended model allows us to estimate the effectiveness of a specific XIL method. As the objective of a XIL method is to overcome the influence of the confounding region, the \wrmeasure score should be at least smaller than the score for the Vanilla model. We expect a XIL method to perform best on its internally-used explainer (i.e., have a lower \wrmeasure score). However, we also inspect the performance of the non-internally-used explainers for comparison to make a more general conclusion about a XIL method.

Besides the overall performance in terms of accuracy, the \wrmeasure score additionally offers informative and reliable insights about the functioning of a XIL method, as they directly quantify the intended optimization objective (e.g., does RRR reduce the IG values in the confounded region?).

\subsection*{D \; Detailed Mathematical Description of XIL Methods}\label{sec:xil_methods_math}
\subsubsection*{D.1 \; Right for the Right Reasons (RRR).}
The objective is to train a differentiable model to be right for the right reason by explicitly penalizing wrong reasons, i.e., irrelevant components in the explanation \cite{Ross2017_RRR}. Here, the explanation $\text{expl}_{\theta}(X)$ is generated with IG. For a model $f(X|\theta) = \hat{y} \in \mathbb{R}^{N \times K}$, IG uses the first-order derivatives $\nabla \hat{y}_n$ w.r.t. its inputs $X \in \mathbb{R}^{N \times D}$. The authors constrain the gradient explanations (IG) by shrinking irrelevant gradients, leading to
\begin{equation}
L_{\text{exp}} = \sum_{n=1}^N \sum_{d=1}^D \left(M_{nd} \frac{\partial}{\partial x_{nd}}\sum_{k=1}^K \log(\hat{y}_{nk})\right)^2
\end{equation}
The loss prevents the model from focusing on the masked region by penalizing large IG values in the region. According to the authors, $L_{\text{pred}}$ and $L_{\text{expl}}$ should have the same order of magnitude by setting a suitable regularization rate $\lambda$. The user's explanation feedback $M$ is propagated back to the model in the optimization phase.

\subsubsection*{D.2 \; Right for the Right Reason GradCAM (RRR-G).}
\cite{Schramowski2020_Plantphenotyping} propose RRR-G as a modification of RRR to utilize the features of a CNN more efficiently. Instead of providing explanations with IG --regularizing the gradients w.r.t. the input-- the authors apply GradCAM as an explanation method to the final convolutional layer to extract a class activation map. The resulting loss is 
\begin{equation}
L_{\text{exp}} = \sum_{n=1}^N \left(M_{n} \, \text{norm}(\text{GradCAM}_{\theta}(X_{n}))\right)^2 \quad ,
\end{equation}
where $\textit{norm}$ normalizes the GradCAM map.

\subsubsection*{D.3 \; Right for the Better Reasons (RBR).}
\cite{Shao2021_RBR} generalize RRR to correct the model's behavior more effectively by using and constraining the model's influence functions (IF) instead of IG. The simplified IF of a training instance $X$ to $\theta$ is based on the second-order derivative of the loss, defined as $I(X, \theta)_{I\!F}^T := I_n \nabla_z \nabla_{\theta} L(X, \hat{\theta})$. $I_n$ denotes the identity matrix, and $\nabla_z \nabla_{\theta} L(X, \hat{\theta})$ gives us the direction for the most significant model change by perturbing a training example $X$. $z$ is the variable to describe the dimensions of $X$. The RBR loss is given by
\begin{equation}
L_{\text{exp}} = \sum_{n=1}^N \sum_{d=1}^D \left(M_{nd} I(X_n,\theta)_{I\!F_{nd}}^T I_{I\!G_{nd}}\right)^2
\end{equation}
According to the authors, this leads to faster convergence, improved quality of the explanations, and improved adversarial robustness of the network, compared to RRR.

\begin{table*}[t]
    \centering
    %\resizebox{\columnwidth}{!}{
    
    \begin{tabular}{cl|l}
        \multicolumn{1}{c}{} & \multicolumn{2}{c}{DecoyMNIST} \\
        XIL & train & test \\ \hline 
        w/o decoy & $99.8 {\scriptstyle\pm0.1}$ & $98.8 {\scriptstyle\pm0.1}$ \\ \hline
        Vanilla & $99.9 {\scriptstyle\pm0.0}$ & $78.9 {\scriptstyle\pm1.1}$ \\ \hline
        RRR & $99.9 {\scriptstyle\pm0.1}$ & $98.8 {\scriptstyle\pm0.1}$ \\
        RRR-G & $99.7 {\scriptstyle\pm0.2}$ & $97.4 {\scriptstyle\pm0.7}$ \\
        RBR & $\mathbf{100}  {\scriptstyle\pm0.0}$ & $\mathbf{99.1} {\scriptstyle\pm0.1}$  \\
        CDEP & $99.3 {\scriptstyle\pm0.0}$ & $97.1 {\scriptstyle\pm0.7}$ \\
        HINT & $97.6 {\scriptstyle\pm0.3}$ & $96.6 {\scriptstyle\pm0.4}$ \\
        CE & $99.9 {\scriptstyle\pm0.0}$ & $98.9 {\scriptstyle\pm0.2}$ \\ 
    \end{tabular}
    \quad
    \begin{tabular}{l|l}
        \multicolumn{2}{c}{DecoyFMNIST} \\
        train & test \\ \hline
        $98.7 {\scriptstyle\pm0.3}$ & $89.1 {\scriptstyle\pm0.5}$ \\ \hline
        $99.5 {\scriptstyle\pm0.2}$ & $58.3{\scriptstyle\pm2.5}$\\ \hline
        $98.7 {\scriptstyle\pm0.3}$ & $\mathbf{89.4} {\scriptstyle\pm0.4}$ \\ 
        $90.2 {\scriptstyle\pm1.6}$ & $78.6 {\scriptstyle\pm4.0}$ \\ 
        $96.6 {\scriptstyle\pm2.3}$ & $87.6 {\scriptstyle\pm0.8}$ \\ 
        $89.8 {\scriptstyle\pm2.7}$ & $76.7 {\scriptstyle\pm3.5}$  \\ 
        $99.0 {\scriptstyle\pm0.9}$ & $58.2 {\scriptstyle\pm2.3}$ \\ 
        $\mathbf{99.1} {\scriptstyle\pm0.2}$ & $87.7 {\scriptstyle\pm0.8}$ \\
    \end{tabular}
    \quad
    \begin{tabular}{l|l|l}
        \multicolumn{3}{c}{ISIC19 } \\
        train & test-P & test-NP\\
        \hline
        -- & -- & -- \\
        \hline
        $100{\scriptstyle\pm0.0}$ & $93.0{\scriptstyle\pm0.3}$ & $88.4{\scriptstyle\pm0.5}$ \\
        \hline
        $\mathbf{100}{\scriptstyle\pm0.0}$ & $94.2{\scriptstyle\pm0.3}$ & $88.1{\scriptstyle\pm0.4}$ \\
        $\mathbf{100}{\scriptstyle\pm0.0}$ & $92.3{\scriptstyle\pm0.3}$ & $\mathbf{88.4}{\scriptstyle\pm0.5}$\\
        $92.6{\scriptstyle\pm5.3}$ & $\mathbf{94.5}{\scriptstyle\pm1.4}$ & $80.3{\scriptstyle\pm5.6}$ \\
        $\mathbf{100}{\scriptstyle\pm0.0}$ & $92.3{\scriptstyle\pm0.3}$ & $87.9{\scriptstyle\pm0.5}$ \\
        $\mathbf{100}{\scriptstyle\pm0.0}$ & $92.8{\scriptstyle\pm0.2}$ & $87.7{\scriptstyle\pm0.5}$ \\
        $\mathbf{100}{\scriptstyle\pm0.0}$ & $92.6{\scriptstyle\pm0.2}$ & $87.5{\scriptstyle\pm0.5}$ \\
    \end{tabular}
    \caption{Mean accuracy scores [\%]. The first row shows performance on the dataset without decoy squares (not available for ISIC). The next row shows the Vanilla model (no XIL) gets fooled, indicated by low test acc. Except for HINT on FMNIST, all methods can recover test accuracy. On ISIC no improvement in terms of accuracy. Best values are bold; cross-validated on 5 runs with std.}
    \label{tab:exp_performance_stds}
\end{table*}

\subsubsection*{D.4 \; Contextual Decomposition Explanation Penalization (CDEP).}
Compared to the other approaches. CDEP \cite{Rieger2019_CDEP} allows the user to directly penalize the importance of certain features and feature interactions using CD as the explanation method $\text{expl}_{\theta}(X)$. Designed for differentiable NNs and given a group of features $\{X_j\}_{j\in S}$, the CD algorithm $g^{CD}(X)$ decomposes the logits $g(X)$ into a sum of two terms, the importance score of the feature group $\beta(X)$ and other contributions $\gamma(X)$ not part of $\beta(X)$. CD is calculated iteratively for all layers of the NN. In contrast to the previous methods, CDEP uses feature groups instead of annotation masks. Given feature groups $X_{i,S},X_i \in \mathbb{R}^D \subseteq \{1, ...,d\}$ for all inputs $X_i$ and the user explanations $\text{expl}_X$, the authors calculate a vector $\beta(X_j)$ for any subset of features $S$ in an input $X_j$. Penalizing this vector leads to

\begin{equation}
L_{\text{exp}} = \sum_{n=1}^N \sum_{S} \left\|\beta(X_{i,S}) - \text{expl}_{X_i, S}\right\|_1
\end{equation}

\subsubsection*{D.5 \; Human Importance-aware Network Tuning (HINT).}
In contrast to previous methods, HINT \cite{Selvaraju2019_HINT} explicitly teaches a model to focus \textit{right reasons} instead of \textit{not} focusing on \textit{wrong reasons}. In other words, HINT rewards activation in regions on which to base the prediction, whereas the previous methods penalize activation in regions on which to \textit{not} base the prediction.
% To realize this, HINT enforces a ranking loss between user annotation mask and GradCAM explanations of the model. The loss aligns the two rankings via a variant of the Weighted Approximate Rank Pairwise (WARP) loss, which is then applied on top of the target model's learning algorithm.
% We use \textit{Simplified HINT} to make it applicable to image classification. Instead of $k$ region proposals per image, we only have whole images as input. 
Here, the user annotation mask $M$ indicates the relevant components with $1$s. To calculate $L_{\text{exp}}$, the mean squared error between the network importance score and the user annotation mask is computed. We calculate the network importance score as the normalized GradCAM value of the input, resulting in the following formulation
\begin{equation}
L_{\text{exp}} = \frac{1}{N} \sum_{n=1}^N \left( \text{norm}(\text{GradCAM}_{\theta}(X_n)) - M_n \right)^2
\end{equation}

\subsection*{E \; Computational Resources}
In the following, we listed some findings of the XIL methods concerning resource efficiency: 
\begin{itemize}

    \item RBR had the longest run times, which can be traced back to the calculation of the second-order derivative, making it infeasible for larger CV datasets.
    
    \item CDEP had the highest requirements in terms of GPU capacity.
    
    \item RRR-G and HINT were very resource-efficient concerning run times and GPU capacity, which can be traced back to the GradCAM explainer method --making it applicable to very deep neural networks.
    
\end{itemize}

\subsection*{F \; Further Experimental Results}\label{sec:further_results}
Additionally, for the ISIC19 dataset, we show two test performances in Tab.~\ref{tab:exp_performance_stds}.
We construct these two different test sets, \textit{test-P} and \textit{test-NP}, as not every image contains the known confounder. ``NP'' means no confounding patch, while ``P'' means at least one confounding patch in the benign test image. This split helps to check in more detail whether the confounding patches lead to differences in performance and are used as informative features for disease classification. It grants additional insights into the train data set. The malignant images in both test sets are identical. We can see that the models achieve higher test-P performances, showing that the confounder is still used to some extent as an informative feature.

In Tab.~ \ref{tab:exp_performance_stds} and Tab.~\ref{tab:feedback_stds}, we show our performance results for (Q1) and (Q2) with the respective standard deviations. Tab.~\ref{tab:feedback_stds} contains an experiment with an additional feedback type, \textit{empty} feedback. It describes a feedback mask that is all-zero, i.e. no user feedback is given. Empty feedback represents another sanity check, to investigate how \textit{no feedback} influences each XIL method as it is expected to not change the model performance compared to the confounded Vanilla model. However, the table shows, that CDEP, once again, improves the model performance, although empty feedback is given. With this, we motivate future research to take a closer look into this explainer and thereby XIL method.

\begin{figure*}[t]
  \centering
  \subfloat[IG]{
  \includegraphics[width=0.59\textwidth]{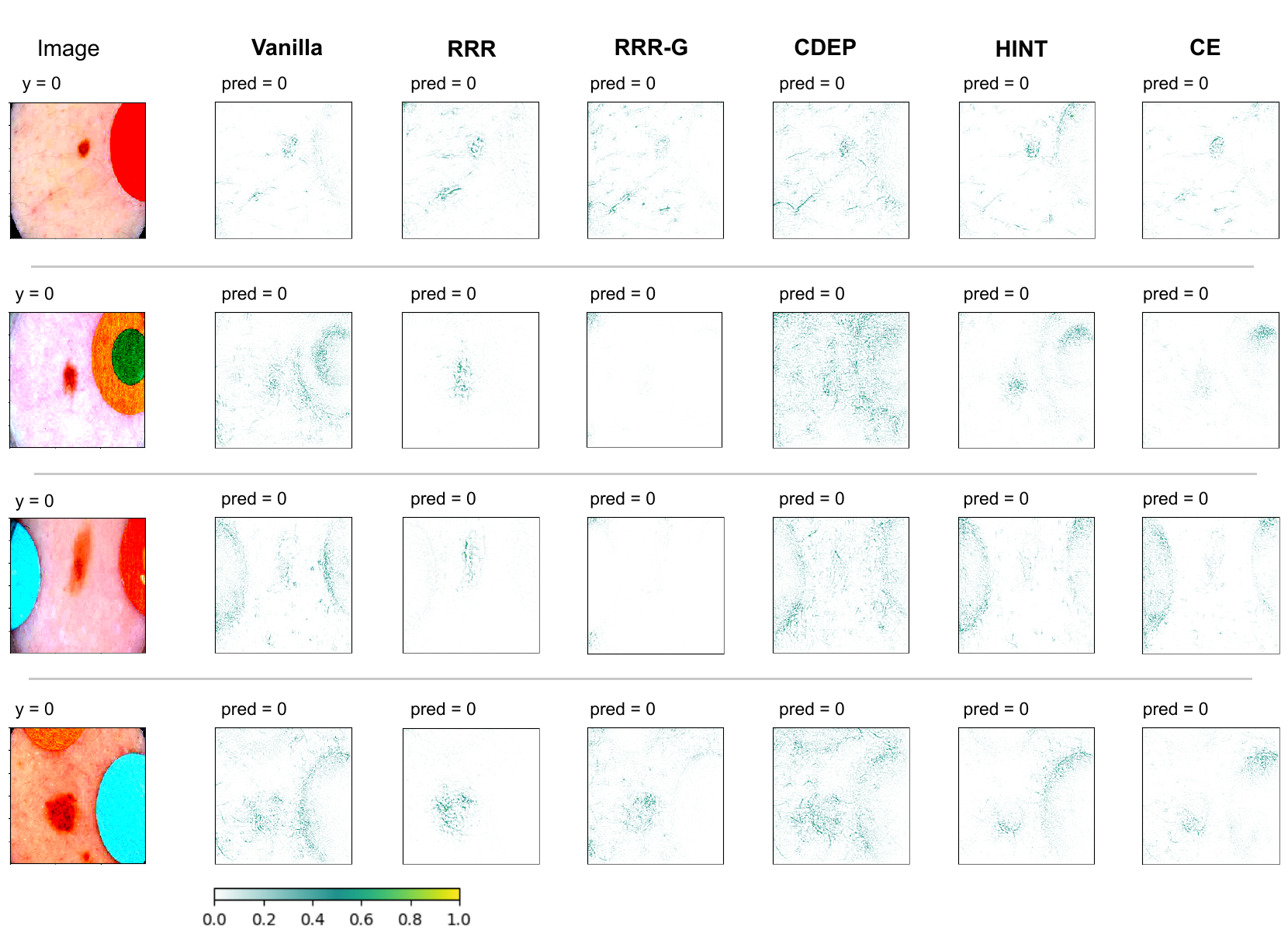}}
%   \caption{ISIC19 exemplary attribution maps generated with IG for the qualitative evaluation.}
%   \label{fig:heatmaps_isic_ig}
% \end{figure*}

% \begin{figure*}[t]
%   \centering
  \subfloat[GradCAM]{
  \includegraphics[width=0.59\textwidth]{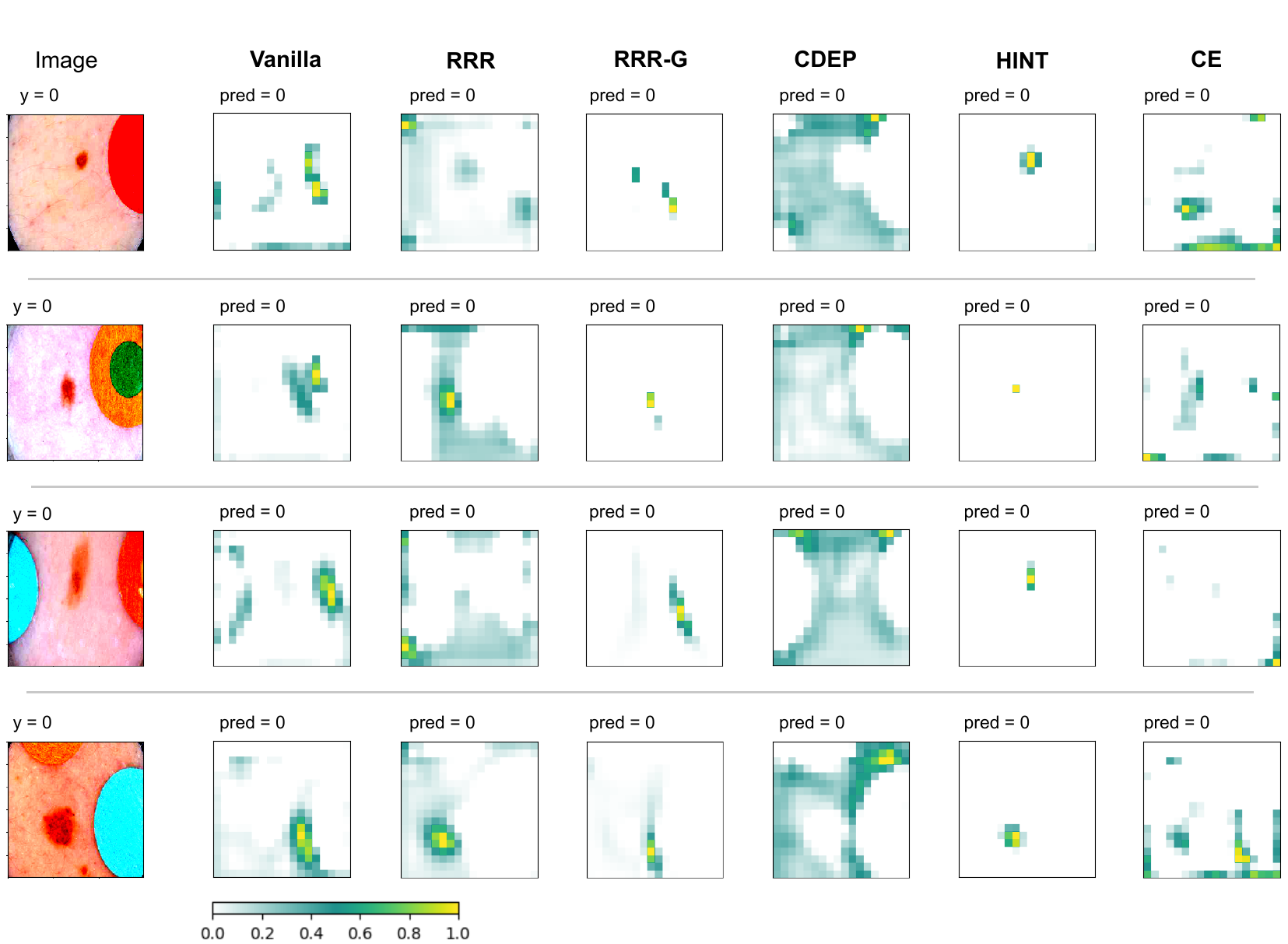}}
%   \caption{ISIC19 exemplary attribution maps generated with GradCAM for the qualitative evaluation.}
%   \label{fig:heatmaps_isic_grad}
% \end{figure*}

% \begin{figure*}[t]
%   \centering
  \subfloat[LIME]{
  \includegraphics[width=0.59\textwidth]{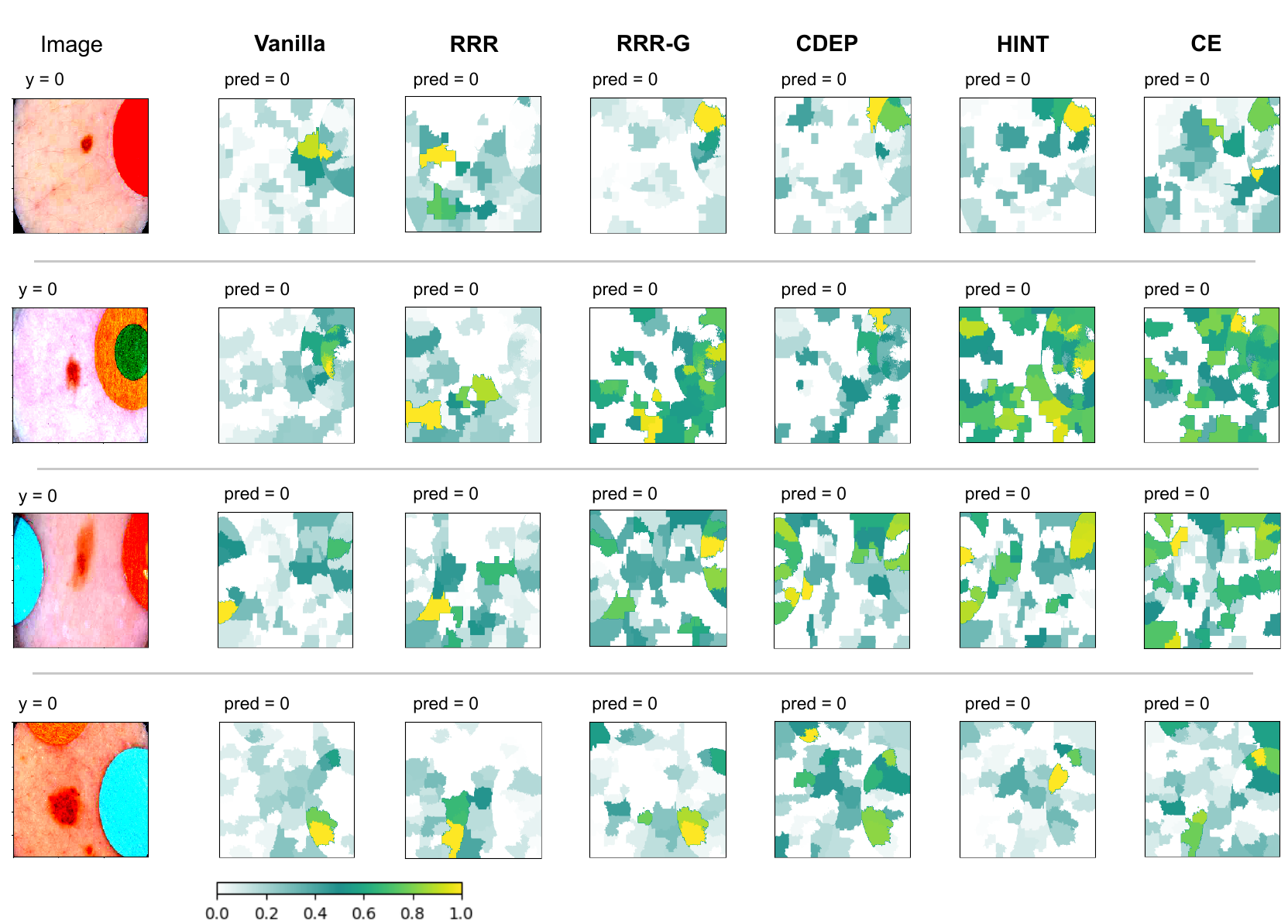}}
  \caption{ISIC19 exemplary attribution maps generated with (a) IG, (b) GradCAM and (c) LIME for a qualitative evaluation.}
  \label{fig:heatmaps_isic}
\end{figure*}

In Fig.~\ref{fig:heatmaps_isic}, we show further qualitative results for the ISIC19 dataset. Each figure uses a different explainer method to visualize the explanation (IG, GradCAM, or LIME).

\begin{table*}[h!p]
    \centering
    
    \def\arraystretch{1}\tabcolsep=2.pt    
    \begin{tabular}[t]{c|c|c|c|c|c|c}
        & \multicolumn{6}{c}{DecoyMNIST (no feedback: $78.9{\scriptstyle\pm 1.1\%}$)} \\
        feedback & RRR & RRR-G & RBR & CDEP & HINT & CE \\ \hline
        arbitrary -- & +$3.3{\scriptstyle\pm6.4}$ & --$4.2{\scriptstyle\pm6.1}$ & --$22.1{\scriptstyle\pm39.6}$ & +$17.8{\scriptstyle\pm1.7}$ & +$4.1{\scriptstyle\pm9.0}$ & +$0.3{\scriptstyle\pm3.0}$ \\
        empty -- & +$1.0{\scriptstyle\pm2.8}$ & +$1.8{\scriptstyle\pm3.0}$ & +$1.5{\scriptstyle\pm1.9}$ & +$3.5{\scriptstyle\pm4.5}$ & --$0.4{\scriptstyle\pm6.3}$ & +$0.2{\scriptstyle\pm1.3}$ \\
        incomplete $\uparrow$ & +$19.6{\scriptstyle\pm1.4}$ & +$9.5{\scriptstyle\pm7.0}$ & +$17.2{\scriptstyle\pm1.8}$ & +$17.9{\scriptstyle\pm1.4}$ & +$17.7{\scriptstyle\pm1.6}$ & +$6.7{\scriptstyle\pm3.2}$ \\
        correct $\uparrow$ & +$19.9{\scriptstyle\pm1.2}$ & +$18.5{\scriptstyle\pm1.8}$ & +$20.2{\scriptstyle\pm1.2}$ & +$18.2{\scriptstyle\pm1.8}$ & +$17.7{\scriptstyle\pm1.5}$ & +$20.0{\scriptstyle\pm1.3}$ 
        \\ \vspace{2pt} \\
        & \multicolumn{6}{c}{DecoyFMNIST (no feedback: $58.3{\scriptstyle\pm 2.5\%}$)}\\
        \hline
        arbitrary -- & +$1.4{\scriptstyle\pm4.4}$ & +$7.9{\scriptstyle\pm4.4}$& --$37.3{\scriptstyle\pm24.5}$ & +$12.2{\scriptstyle\pm4.8}$& --$3.2{\scriptstyle\pm18.2}$& --$1.2{\scriptstyle\pm3.9}$ \\
        empty -- & +$1.3{\scriptstyle\pm3.9}$ & +$0.5{\scriptstyle\pm4.0}$& +$1.4{\scriptstyle\pm3.9}$ & +$6.0{\scriptstyle\pm8.7}$& --$2.3{\scriptstyle\pm4.7}$& --$0.4{\scriptstyle\pm3.1}$ \\
        incomplete $\uparrow$ & +$24.2{\scriptstyle\pm4.0}$ &+$12.4{\scriptstyle\pm5.1}$ & +$16{\scriptstyle\pm4.0}$ & +$16.8{\scriptstyle\pm7.2}$& +$21{\scriptstyle\pm4.4}$& +$3.4{\scriptstyle\pm3.9 }$ \\
        correct $\uparrow$ & +$31.1{\scriptstyle\pm2.9}$ &+$20.3{\scriptstyle\pm4.2}$ & +$29.3{\scriptstyle\pm3.3}$ &+$18.4{\scriptstyle\pm2.6}$ & --$0.1{\scriptstyle\pm4.8}$ & +$29.4 {\scriptstyle\pm3.3}$\\

        % arbitrary -- & +$5.4 $ & --$4.5 $ & +$7.5 $ & +$14.1 $ & --$9.0 $ & +$0.0 $ & --$1.6 $ & & +$6.6 $ & & & +$1.0 $ \\ 
        % incomplete $\uparrow$ & +$19.7 $ & +$15.0 $ & +$18.2 $ & +$17.7 $ & +$17.9 $ & +$7.3 $ & +$26.1 $ & & +$16.4 $ & & & +$4.3 $ \\
        % correct $\uparrow$ & +$19.9 $ & +$18.5 $ & +$20.2 $ & +$18.2 $ & +$17.7 $ & +$20.0 $ & +$31.1 $ & & +$25.1 $ & & & +$29.4 $\\
    \end{tabular}
    \caption{Feedback robustness evaluation on Decoy(F)MNIST for arbitrary, empty, and incomplete compared to correct feedback masks. The mean difference in test accuracy [\%] compared to the confounded Vanilla model is given; cross-validated on 5 runs with std. For arbitrary and empty feedback, unchanged is better, and for incomplete and correct feedback, higher is better. Incomplete feedback is on par with correct feedback.}
    \label{tab:feedback_stds}
\end{table*}  

% \bibliographystyle{bst/sn-vancouver}
% \bibliography{references}

% \end{document}

\end{document}